%% file: main.tex
\documentclass[10pt,twocolumn,letterpaper,table]{article}

\usepackage[pagenumbers]{cvpr}      

\input{preamble}
\definecolor{cvprblue}{rgb}{0.21,0.49,0.74}
\usepackage{hyperref}
\hypersetup{pagebackref,breaklinks,colorlinks,allcolors=cvprblue}

\hypersetup{
    linkcolor=cvprblue,
    citecolor=NavyBlue!50!Gray,
}

\def\paperID{} 
\def\confName{CVPR}
\def\confYear{2025}

\title{
  DINOv2 Meets Text: A Unified Framework for Image- and Pixel-Level Vision-Language Alignment
}

\author{
Cijo Jose$^1$ \hspace{1em}
Théo Moutakanni$^{1,2}$ \hspace{1em}
Dahyun Kang$^{1,3,*}$ \hspace{1em}
Federico Baldassarre$^1$  \\
Timothée Darcet$^{1,4}$  \hspace{1em}
Hu Xu$^1$ \hspace{1em}
Daniel Li$^1$ \hspace{1em}
Marc Szafraniec$^1$ \hspace{1em}
Micha\"el Ramamonjisoa$^1$ \\
Maxime Oquab$^1$ \hspace{1em}
Oriane Siméoni$^1$ \hspace{1em}
Huy V. Vo$^1$ \hspace{1em}
Patrick Labatut$^1$ \hspace{1em}
Piotr Bojanowski$^1$ \\ \\
\normalsize{$^1$ Meta FAIR\quad $^2$ Université Paris-Saclay, CentraleSupélec, MICS \quad} \\ \normalsize{$^3$ POSTECH\quad $^4$ Univ. Grenoble Alpes, Inria, CNRS, Grenoble INP}
}
\vspace{-10pt}

\begin{document}

\maketitle

\begin{abstract}
Self-supervised visual foundation models produce powerful embeddings that achieve remarkable performance on a wide range of downstream tasks. However, unlike vision-language models such as CLIP~\cite{radford2021learning}, self-supervised visual features are not readily aligned with language, hindering their adoption in open-vocabulary tasks. Our method, named \oursnospace, unlocks this new ability for DINOv2~\cite{oquab2024dinov2}, a widely used self-supervised visual encoder.
We build upon the LiT training strategy~\cite{zhai2022lit}, which trains a text encoder to align with a frozen vision model but leads to unsatisfactory results on dense tasks. We propose several key ingredients to improve performance on both global and dense tasks,
such as concatenating the \cls\xspace token with the patch average to train the alignment and curating data using both text and image modalities. With these, we successfully train a CLIP-like model with only a fraction of the computational cost compared to CLIP while achieving state-of-the-art results in zero-shot classification and open-vocabulary semantic segmentation.
\end{abstract}

\vspace{-10pt}
\section{Introduction}
\blfootnote{$^*$: work done during an internship at Meta.}
The advent of modern vision foundation models trained in a Self-Supervised Learning (SSL) fashion
\cite{oquab2024dinov2, he2021masked, chen2021empirical, assran2023self, caron2021emerging}
has resulted in robust, generic features that achieve impressive performance on downstream tasks.
These features are typically used \textit{as is}, plugged into a light-weight adapter such as a linear classifier, and deliver strong results without requiring a costly fine-tuning process. As a result, a single strong vision backbone can be used simultaneously for different tasks.
DINOv2~\cite{oquab2024dinov2}, in particular, has been popular for its versatility.
This self-supervised model, trained to capture both the global context and local information of the image, has led to state-of-the-art performance in tasks that require an overall understanding of the image such as classification and those that necessitate more fine-grained details such as segmentation~\citep{li2024promerge}, canopy height prediction~\citep{tolan2023sub}, object matching~\citep{ornek2025foundpose,Edstedt_2024_CVPR}, object discovery~\cite{didolkar2024zero} and tracking~\citep{dino_tracker_2024,zadaianchuk2024object}.
However, self-supervised vision models
do not provide an interface with language, limiting their use in open-vocabulary scenarios in which multi-modal models~\citep{chen2024internvl,sun2024eva} that come with built-in language-vision alignment excel.
This is a notable weakness in the era of complex and promptable machine learning systems.
We aim in this work to equip DINOv2 with a language interface by aligning its feature space with language, which allows us to leverage the strengths of this powerful self-supervised model to tackle open-vocabulary recognition tasks.

\begin{figure*}[ht!]
    \centering

    \begin{minipage}{0.4\textwidth}
    \centering
    \caption*{Zero-shot \textbf{classification}}
    \vspace{-5pt}
    \end{minipage}
    \begin{minipage}{0.55\textwidth}
    \centering
    \caption*{Open-vocab. \textbf{segmentation} (ADE20K classes, please zoom in for details)}
    \vspace{-5pt}
    \end{minipage}

    \begin{minipage}{0.4\textwidth}
    \centering
    \includegraphics[width=0.85\linewidth]{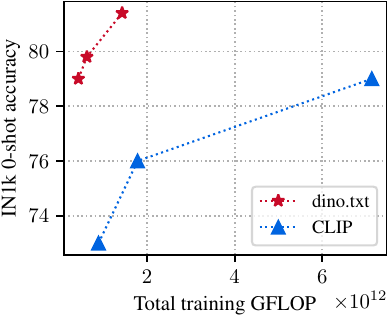}
    \end{minipage}
    \begin{minipage}{0.55\textwidth}
    \centering
    \setlength{\tabcolsep}{1pt}
    \begin{tabular}{cc}
      \includegraphics[width=0.25\linewidth]{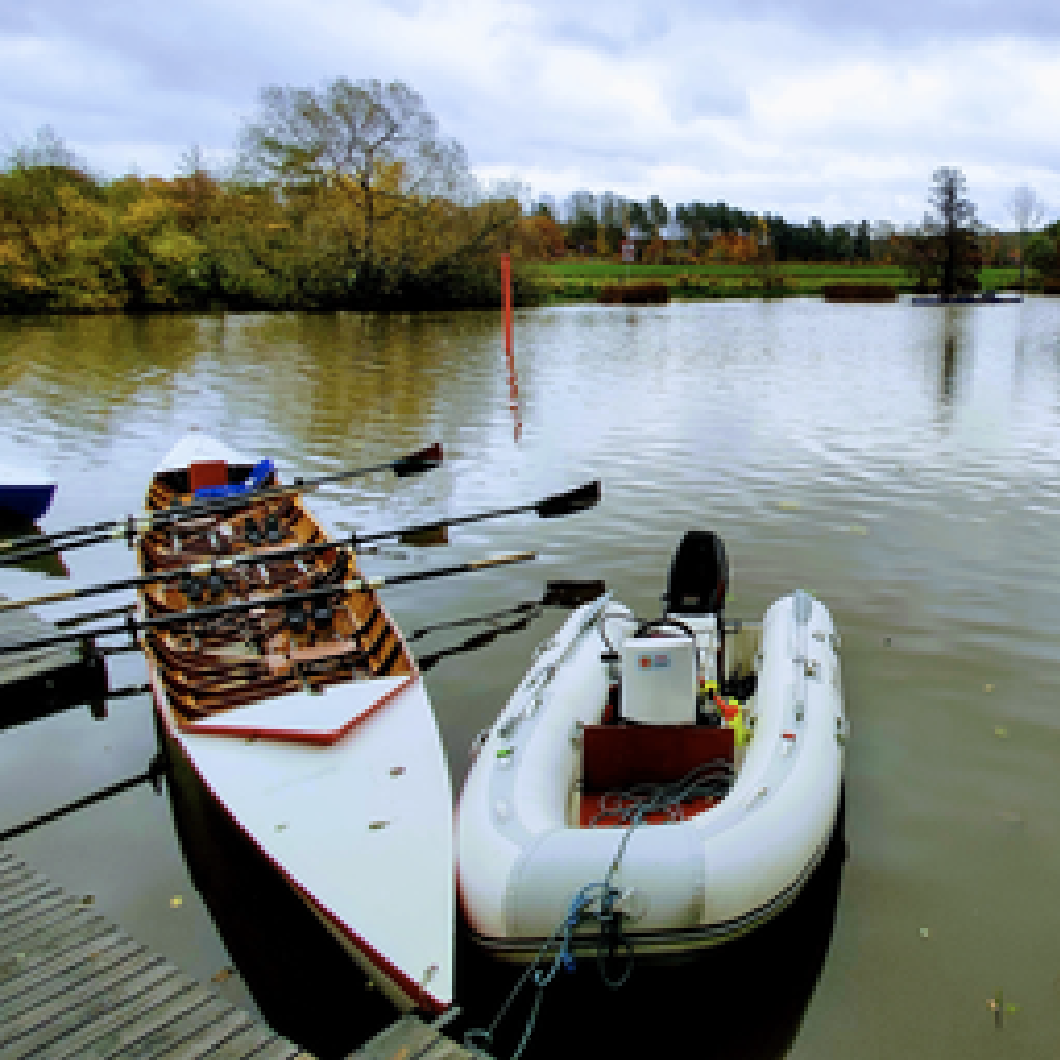}
       \includegraphics[width=0.25\linewidth]{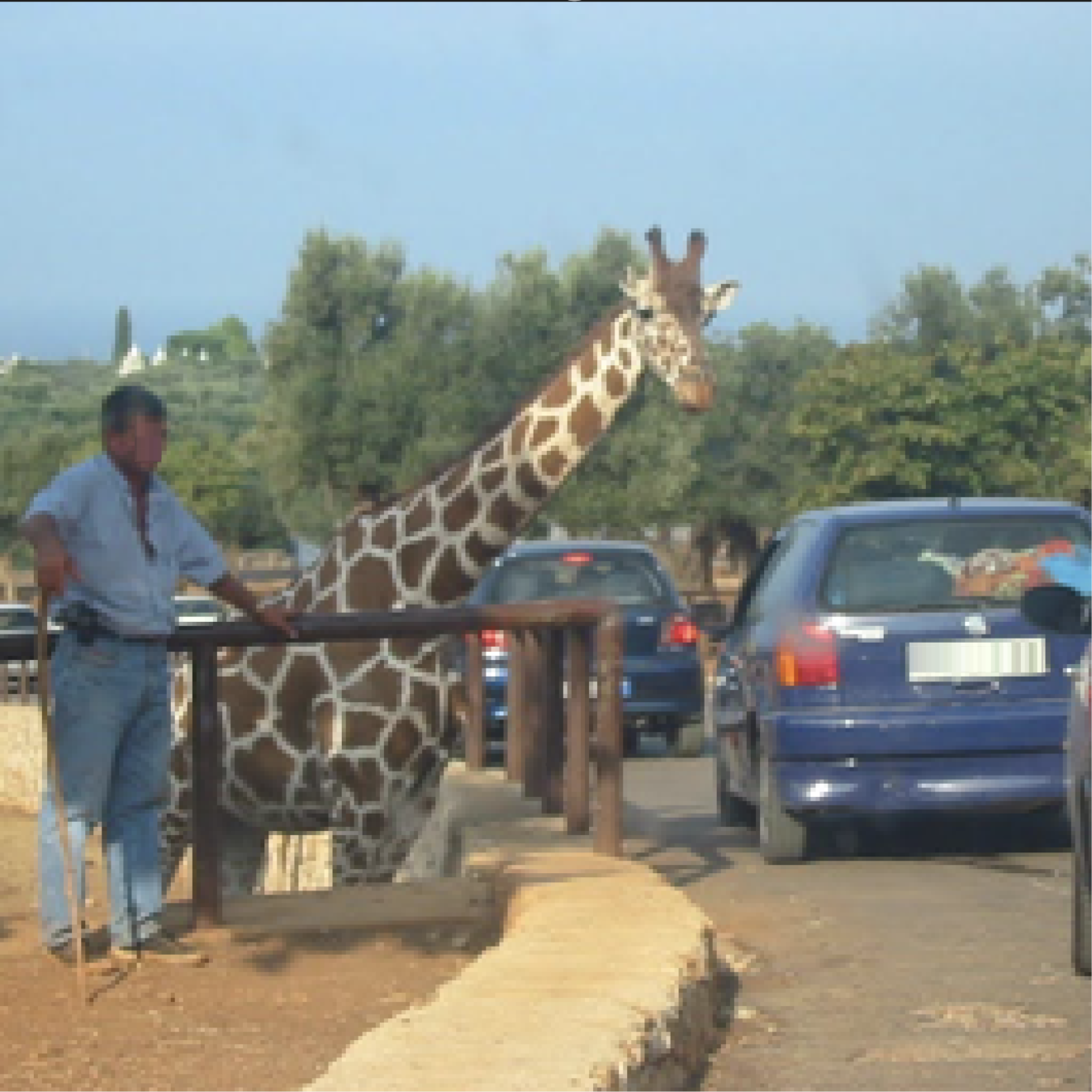}  & \includegraphics[width=0.25\linewidth]{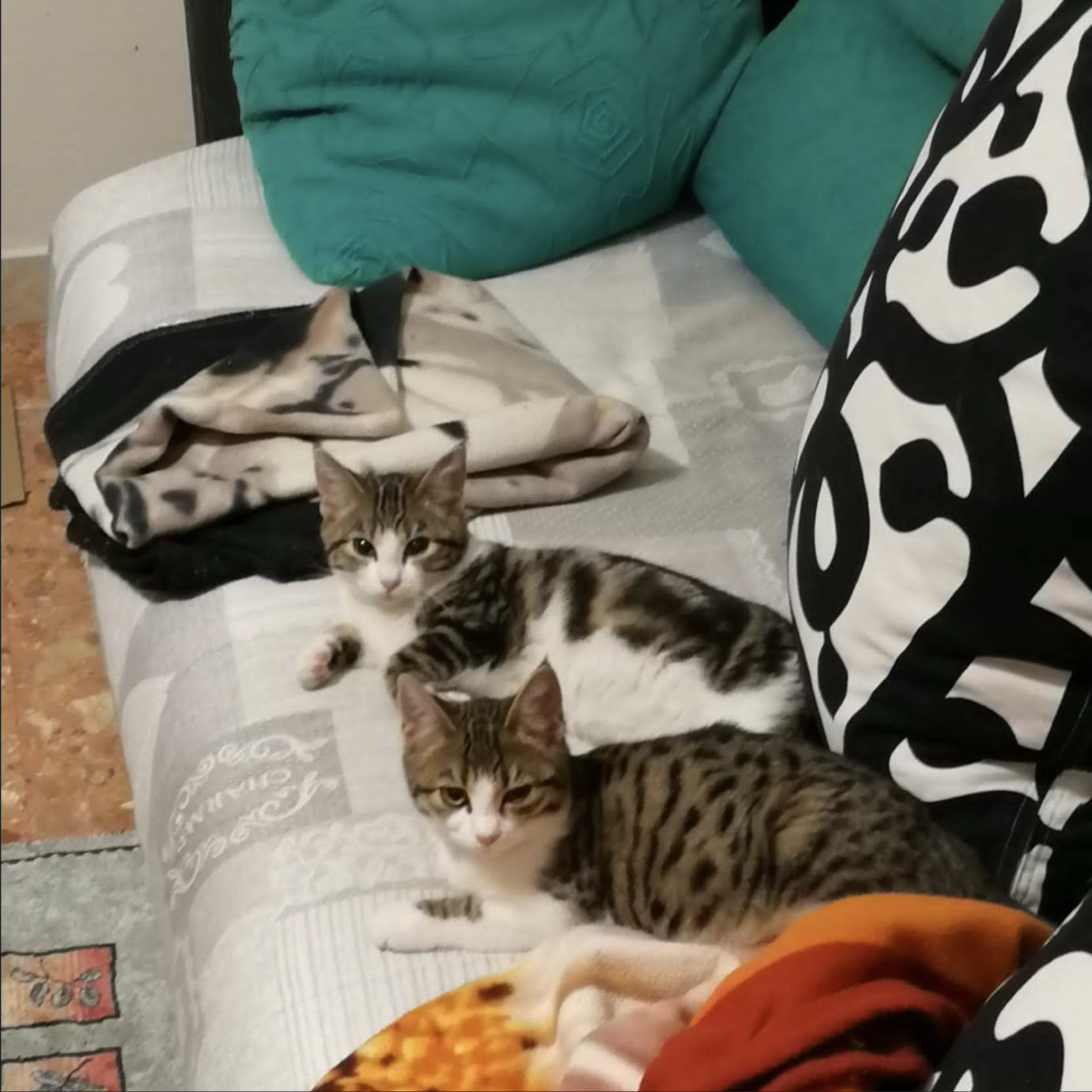}
       \includegraphics[width=0.25\linewidth]{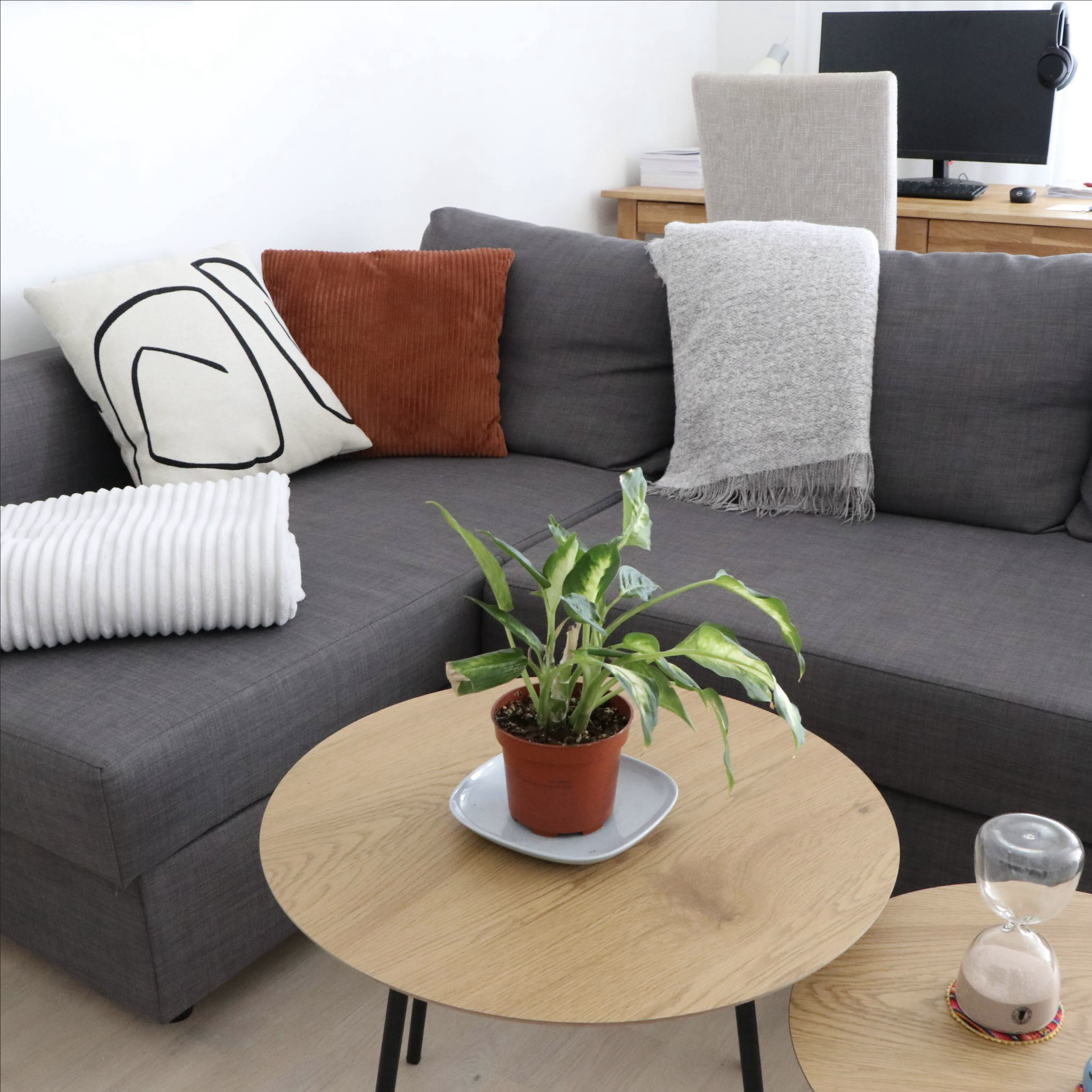}\\
       \includegraphics[width=0.25\linewidth]{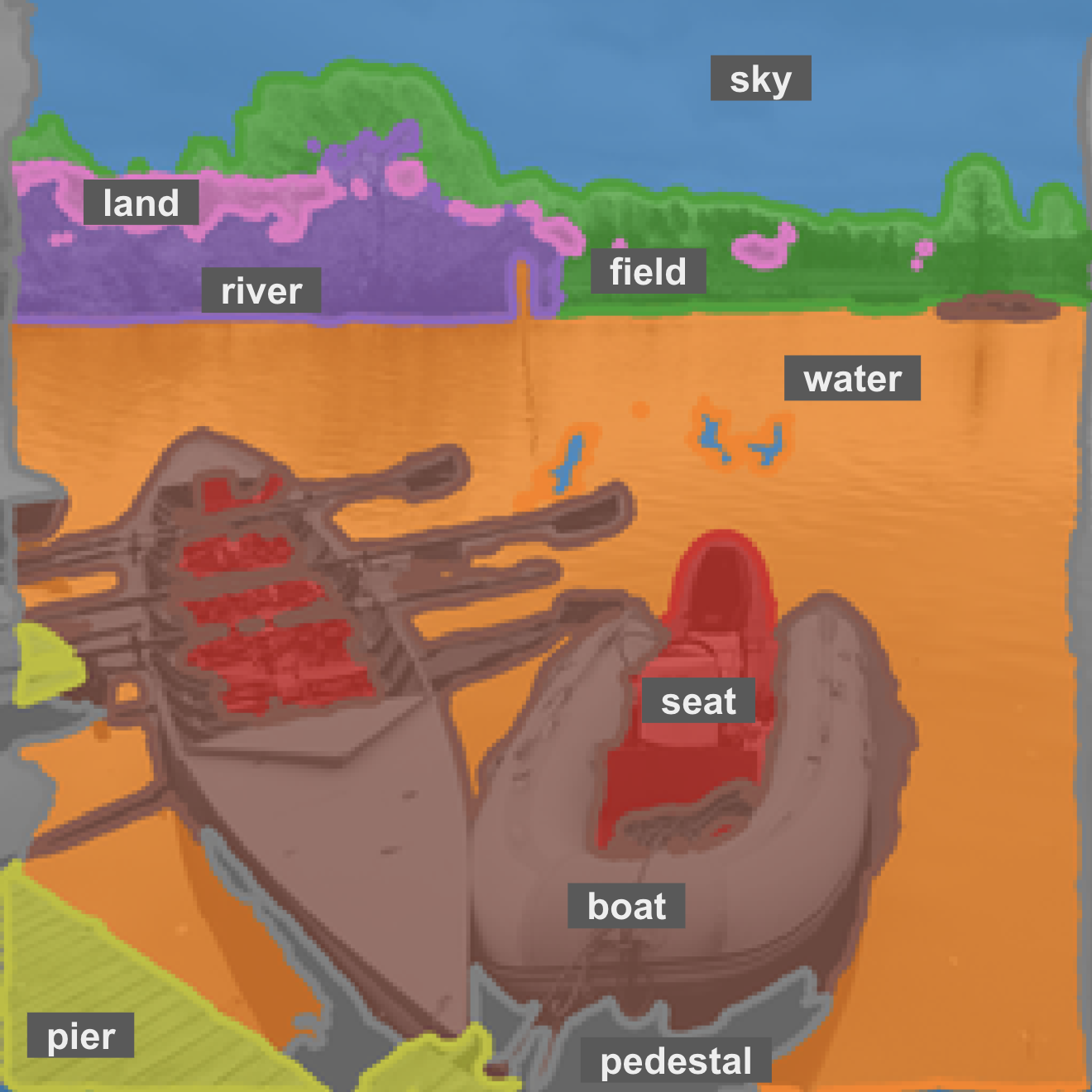}
       \includegraphics[width=0.25\linewidth]{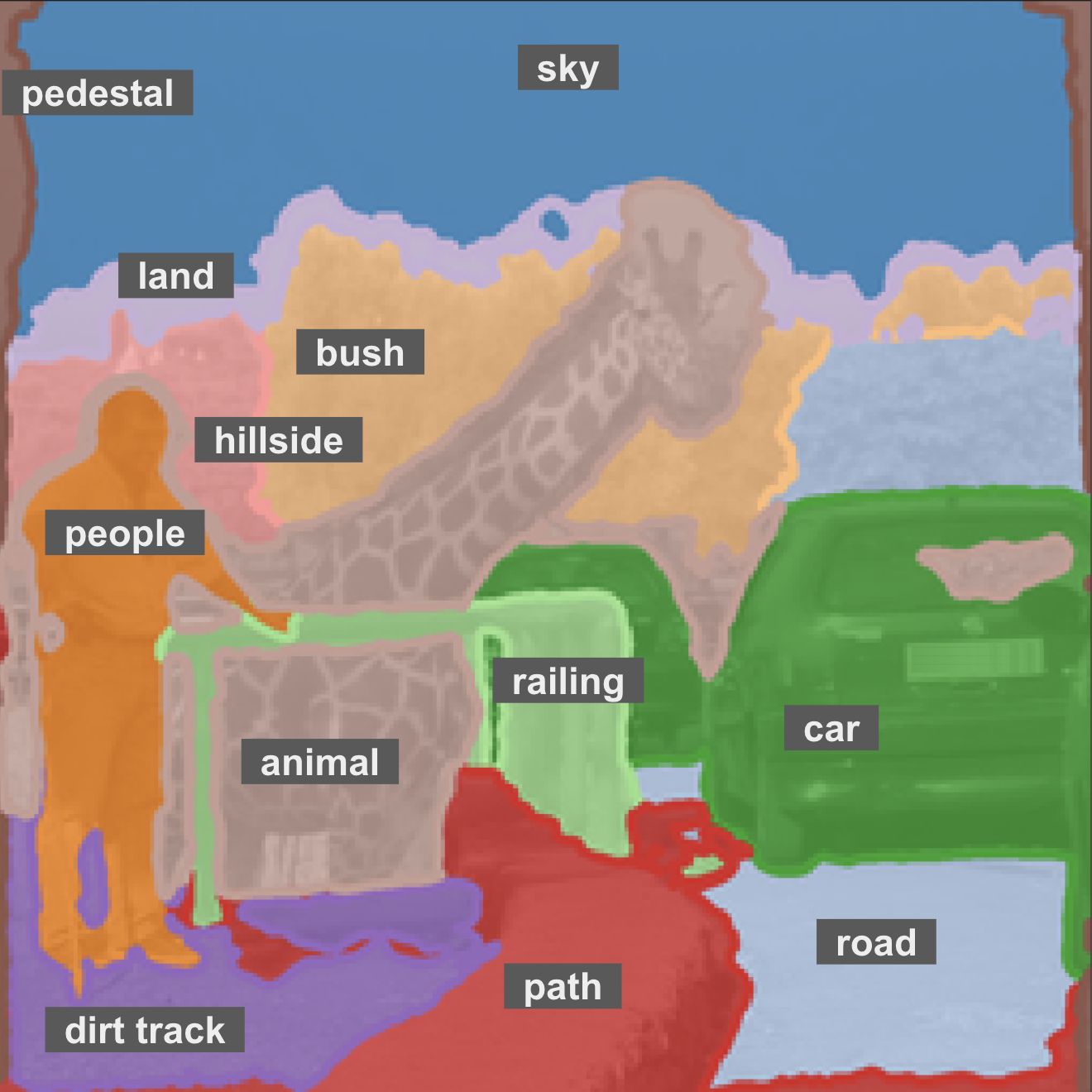}  & \includegraphics[width=0.25\linewidth]{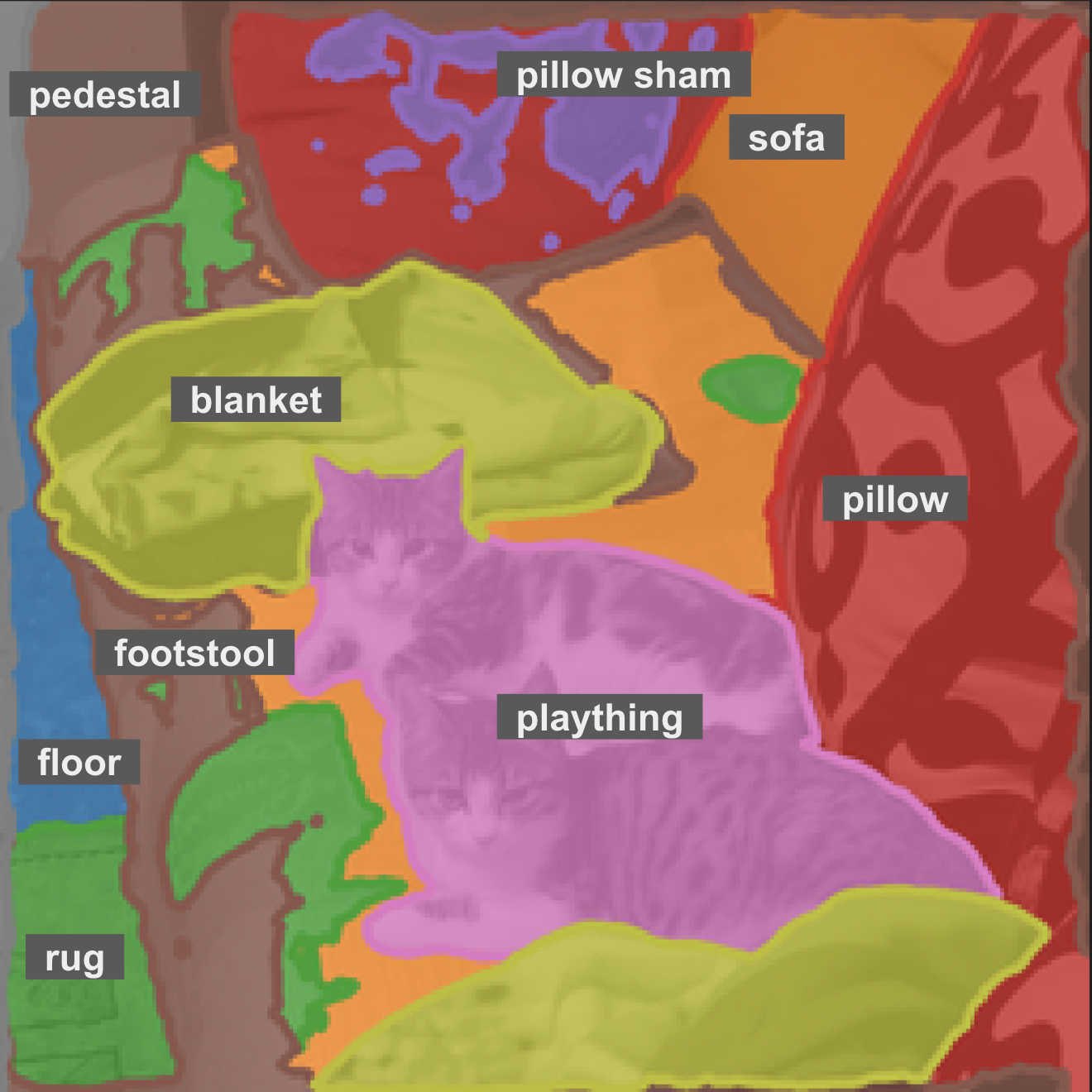}
       \includegraphics[width=0.25\linewidth]{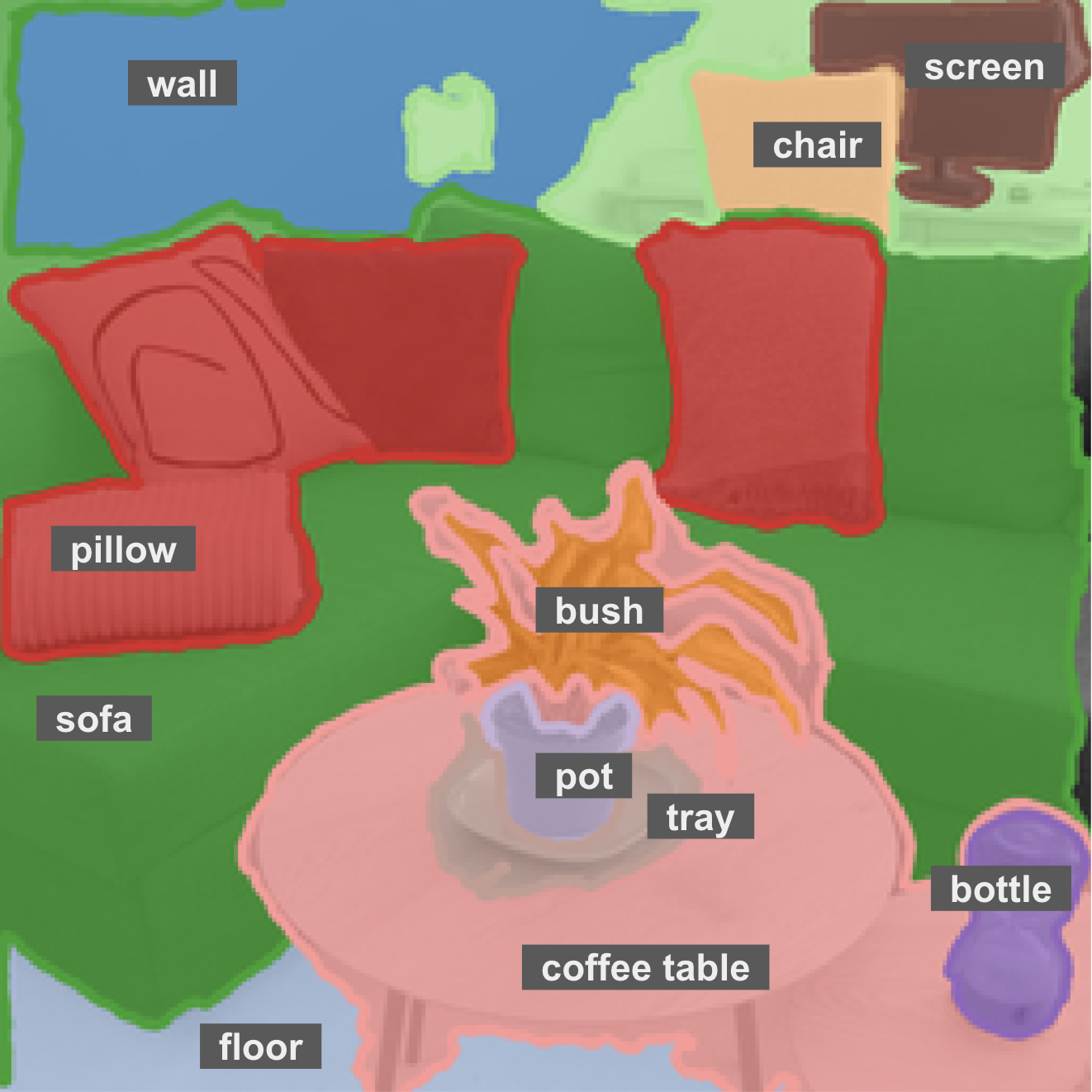}
    \end{tabular}
    \end{minipage}

    \vspace{-10pt}
    \caption{\textbf{Zero-shot classification and open-vocabulary segmentation results with our method \ours trained with only weak image/caption annotation.} We show that our training strategy leads to state-of-the-art results in zero-shot classification with a fast and efficient training. Also, its produces high quality segmentation results on diverse images showing the quality of the image-to-text alignment.}
    \vspace{-2ex}
    \label{fig:intro-fig}
\end{figure*}

Most advanced text-aligned vision foundation models learn with a variation of the CLIP algorithm~\cite{radford2021learning}, which trains the visual and textual encoders to align their modality representation in a shared space by
exploiting a large-scale, often noisy, paired image-text dataset.
They are typically trained from scratch, leading to heavy computational cost.
Locked-image text tuning (LiT)~\citep{zhai2022lit} is a variant of CLIP that uses a frozen pre-trained vision model as the vision encoder and only train the text encoder to align to the vision encoder's embedding space. This leads to a lower computational cost while retaining desirable properties of the vision encoder. In this work, we argue that given readily available strong vision encoders, we could, and should achieve better vision-language alignment than CLIP at a much smaller cost.
To this end, we explore the application of LiT training with DINOv2 as the vision encoder.

As shown in \autoref{tab:ablation:visualfeatures},~\ref{tab:ablation:layers2}, applying LiT on strong DINOv2 encoder is not straightforward as it leads to unsatisfying performance on tasks that require fine-grained details such as semantic segmentation or image-text retrieval.
First, it is not trivial to obtain
good dense features from a model trained with the CLIP/LiT training paradigm, which contrasts only global image and text representation.
Second, the domain gap between visual pre-training data and LiT training data caused by the frozen vision encoder could potentially hinder the alignment of images and their captions.
To address these issues, we introduce several improvements to the LiT paradigm.
Instead of the commonplace practice of using the \cls\xspace token from the vision encoder to represent the image, we concatenate this token to the average pool of all patch tokens in the image as the vision representation to allow aligning both the global context and the local information of the image to its textual description.
Then, we reduce the aforementioned domain gap by adding two
learnable vision blocks on top of the frozen vision encoder, thereby allowing vision features to adapt to the new training data. We show the benefit of our training for zero-shot classification and open-vocabulary segmentation\footnote{We follow the taxonomy in
the survey~\cite{wu2024towards} and use the nomenclature `open-vocabulary' segmentation.} in \autoref{fig:intro-fig}.

Moreover, the quality of pre-training data has been shown to strongly influence
the model's performance~\citep{radford2021learning,xu2024demystifying} but also the training efficiency~\citep{fang2024data}. We show that by paying attention to the dataset curation, we can further improve our training procedure.
Indeed, we propose to curate the training dataset by balancing the long-tailed distribution of image and text data.
A well-balanced data distribution eases the training, allowing us to reach good performance with only a fraction of the computational cost. This in turns allows us to experiment with a wider text encoder, leading to further improvements in performance.
Finally, our study
not only
unlocks text alignment for DINOv2 but also reveals limitations in the LiT framework discussed in the error analysis section,
pointing toward future directions for more effective and efficient frameworks to achieve language-aligned vision foundation models.

To summarize, the contributions of this work include:
\textbf{(i)}
a new method \oursnospace, which unlocks image- and pixel-level text-alignment for DINOv2,
\textbf{(ii)}
key ingredients on top of existing works that allow to train such multi-modal alignment for only a fraction of the usual compute cost,
\textbf{(iii)}
an extensive error analysis, demonstrating the limitations of existing segmentation benchmarks for this task and the different error types with these models.

\input{related}
\input{problem}
\input{experiments}
\input{error_analysis}

\section{Conclusion}
We have presented a training recipe, named \oursnospace, which aligns from scratch a text encoder to a frozen self-supervised vision model, specifically DINOv2 \cite{oquab2024dinov2, darcet2023vision}
, unlocking open-vocabulary abilities.
The approach includes a self-supervised data curation technique with no human annotation and allows for fast training, leading to strong zero-shot classification performance, on par with the state-of-the-art.
The resulting text encoder is also aligned to patch-level features,
therefore providing precise dense open-vocabulary segmentation capabilities thanks to the quality of the frozen vision encoder.
We also argue that classic semantic segmentation benchmarks
require rethinking for open-vocabulary, as they do not allow for overlapping concepts nor finer granularity in prediction than the annotations.

\small
\bibliographystyle{ieeenat_fullname}
\bibliography{main.bib}

\newpage
\clearpage
\newpage
\appendix
\input{supp}

\end{document}

%% file: preamble.tex
\usepackage{multirow}
\usepackage{layouts}
\usepackage{float}
\usepackage{placeins}

\usepackage{amssymb}
\usepackage{pifont}

\usepackage{lscape}
\usepackage{rotating}

\newcommand{\ours}{\texttt{dino.txt}~}
\newcommand{\oursnospace}{\texttt{dino.txt}}

\newcommand{\Reals}{\mathbb{R}}

\newcommand{\venc}{\phi_{\text V}}

\newcommand{\tenc}{\phi_{\text T}}
\newcommand{\eos}{\texttt{[EOS]}}
\newcommand{\cls}{\texttt{[CLS]}}
\newcommand{\image}{\mathbf{I}}
\newcommand{\dataset}{LVTD-2.3B}

\newcommand{\pool}{\texttt{pool}}

\newcommand\blfootnote[1]{
    \begingroup
    \renewcommand\thefootnote{}\footnote{#1}
    \addtocounter{footnote}{-1}
    \endgroup
}

%% file: related.tex
\section{Related work}


\vspace{-5pt}
\paragraph{Self-supervised feature learning.}
Visual features from image encoders trained in a self-supervised fashion have been used in many machine learning systems due to their good performance and generalizability.
Multiple approaches for learning these models have been developed in recent years.
Among these, contrastive learning~\citep{oord2018representation} trains models to pull features of similar images while pushing those of dissimilar images. Notable methods include MoCo~\citep{he2020momentum} which employs a memory bank, BYOL~\citep{grill2020bootstrap} which removes the need for negative pairs, SwAV~\cite{caron2020unsupervised} which contrasts online cluster assignments, or DINO~\cite{caron2021emerging} which extends SwAV to Vision Transformers~\cite{dosovitskiy2020image}.
In contrast, reconstruction-based methods learn by predicting hidden portions of input images such as missing pixels (MAE~\cite{he2021masked}), patches from quantized code book (BeiT~\citep{bao2021beit}) or patch features in a latent space (I-JEPA~\citep{assran2023self}).
With such an array of viable approaches, a determining factor remains whether the SSL methods can improve with increasing data and model sizes: scaling was explored in multiple works~\cite{goyal2019scaling,caron2019unsupervised,goyal2021self,goyal2022vision}, with DINOv2~\cite{oquab2024dinov2} setting the current state of the art for this problem.


\vspace{-10pt}
\paragraph{Contrastive text-image pre-training.}
The idea of leveraging textual metadata to train image understanding models has a long history in computer vision~\cite{duygulu2002object,gupta2008beyond,farhadi2010every,kulkarni2013babytalk}.
In the context of deep neural networks, Joulin \etal~\cite{joulin2016learning} proposed to use words from image captions as targets to train visual encoders.
This core idea has been further improved in CLIP~\cite{radford2021learning}.
The authors propose to encode the image and the caption, and train both using a contrastive loss.
Deep encoding of captions facilitates robust generalization across sentences, such as by collapsing synonyms, thereby enhancing learning efficiency.
Since the original CLIP was trained on a private dataset, open source reproduction attempts have focused on collecting public large-scale  datasets (LAION~\cite{schuhmann2022laion}), leading notably to the OpenCLIP~\cite{cherti2023reproducible} family of models. More recently, \citet{xu2024demystifying} described a simple procedure for re-balancing image-text web data, reproducing the performance of CLIP.
The zero-shot performance was further improved by DFN~\cite{fang2024data} which proposes filtering the training data to match the distribution of downstream tasks.
Even more refined systems have recently been developed: EVA-CLIP~\cite{sun2023eva,sun2024eva} or InternVL~\cite{chen2024internvl},
demonstrated at scales above 5 billions parameters and further narrowed the gap between fully-supervised and zero-shot models on ImageNet.

Apart from data and model scaling, a few modifications to the initial training algorithm have been proposed.
SigLIP~\cite{zhai2023sigmoid} considers a binary log loss instead of multinomial cross entropy. LLiP~\cite{lavoie2024modeling} proposes applying the CLIP loss between the text and image register tokens~\cite{darcet2023vision} that are conditioned on the text tokens.
In contrast, we do not use any improved loss function in our training, and stick to CLIP's original contrastive loss with a frozen image encoder and a learnable text encoder, following the procedure described in locked-image text tuning (LiT~\cite{zhai2022lit}).


\vspace{-10pt}
\paragraph{Automatic data curation at scale.}
Automatic data curation plays a crucial role in the training of foundation models with massive-scale web-crawled datasets, which typically reach over hundreds of millions~\cite{schuhmann2021laion, xu2024demystifying, gadre2024datacomp} of data.
At this scale, manual annotation becomes infeasible, so these datasets are often collected without supervision from the internet.
Such in-the-wild data inherently exhibit a long-tailed distribution of data categories~\cite{Liu2019longtail} which limits a model's ability to efficiently learn to cover broad concepts.
To address this issue, related works on foundation models often construct balanced training datasets by suppressing head (frequent) concepts and boosting tail (rare) ones.
For example, CLIP's~\cite{radford2021learning} data preparation pipeline collects five hundred thousand frequent words and queries each word in the raw dataset to retrieve a balanced number of (image, caption) pairs. The unrevealed details of this pipeline are later reproduced and formalized by MetaCLIP~\cite{xu2024demystifying}.
DFN~\cite{fang2024data} trains a data evaluation model that assesses the ``quality'' of data to sample the top samples among the raw data pool.
SemDeDup~\cite{abbas2023semdedup} prunes it by removing duplicated data points detected with clustering.
Finally, \citet{vo_automatic_2024} balances data distribution by sampling uniformly over the data support.
These methods apply data curation to either images or captions, while we balance both distributions, leading to better performance and more efficient training.


\paragraph{Open-vocabulary segmentation.}
CLIP models can be adapted to produce patch-level features aligned with text by performing several forward passes on different views of the image \citep{abdelreheem2023satr,kerr2023lerf, wysoczanska2023clipdiy, jatavallabhula2023conceptfusion} or producing code books \citep{shin2022reco,shin2023namedmask,karazija2023diffusion} of prototypes per concept of interest. MaskCLIP~\citep{zhou2022maskclip}, which can be applied to most vision-language models (VLMs), adapts the model by removing the final global pooling and applying the final projection to the value embedding of the last attention layer, achieving dense features in CLIP space. Such features can be refined with improved attention mechanisms \citep{bousselham2024grounding,wang2025sclip}, or using an SSL model as a guide \citep{wysoczanska2023clipdino,lan2024proxyclip,kang2024lazy}. Such efforts are orthogonal to our work as they can be applied to any dense CLIP-like features.

\begin{figure*}[ht!]
    \centering
    \includegraphics[trim={0 0 0 0.1cm},clip, width=\linewidth]{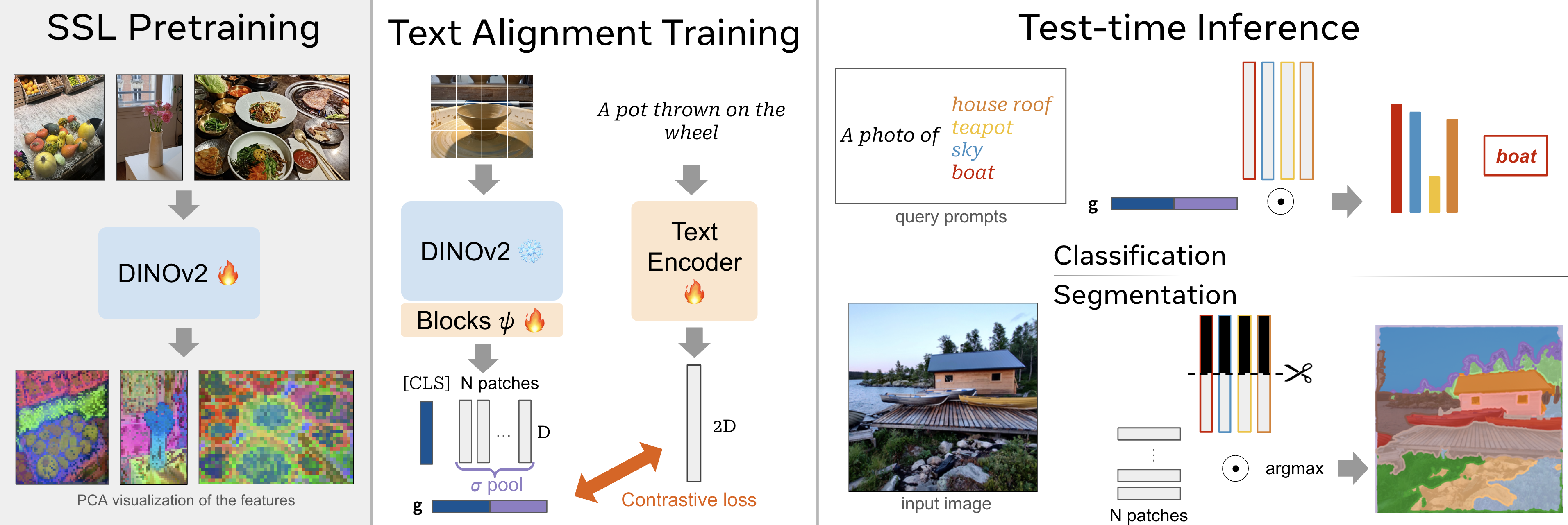}
    \caption{
        \textbf{Overview of our method \oursnospace}. We first show the localization quality of the self-supervised features (left). We then present our training strategy (middle) which consists in aligning the frozen SSL backbone to a text-encoder trained from scratch. We additionally add a light vision block on top of the visual encoder in order to better align with the text. We train our model for just 50k iterations and achieve SoTA results on both zero-shot classification and open-vocabulary segmentation  (right).
    }
    \vspace{-2ex}
    \label{fig:method overview}
\end{figure*}

Improved patch-level alignment can be obtained by fine-tuning or training from scratch a CLIP-like model with dedicated objectives using
pixel-level annotations~\citep{li2022languagedriven, liang2023open,rao2021denseclip, ghiasi2022openseg, ding2021ZegFormer,wang2024sam} or
coarse image/caption annotations
~\citep{ghiasi2022openseg,clippy2022,zhong2022regionclip,lou2022segclip,liang2023open,xu2022groupvit,liu2022open,xu2023learning,mukhoti2023pacl,cha2022tcl}. In this work, we focus on the latter.
ViewCO~\citep{ren2023viewco} leverages multi-view consistency and CLIPSelf~\citep{wu2023clipself} uses a teacher-student learning strategy to produce dense features aligned with those obtained from crops.
GroupViT \citep{xu2022groupvit} integrates learnable tokens that are trained using grouping blocks,
which CoCu~\cite{NEURIPS2023_d937cb3f} further improves by using image retrieval on image-caption pairs to create concept banks and use them as training data.
PACL~\citep{mukhoti2023pacl} trains patch-to-text affinity with a dedicated module, TCL~\citep{cha2022tcl} proposes a local contrastive objective to align well-selected patches to the text, and
CoDe~\cite{wu2024imagetextcodecompositiontextsupervisedsemantic} uses a word-region local contrastive objective to match regions of the image to segments of the text.
Closer to us, CLIPpy~\cite{clippy2022} fine-tunes an SSL vision backbone and pre-trained text encoder to produce aligned features (using the average patch), however, at the cost of worse classification results. In this work, we show that it is possible to train a model with both image- and pixel-level alignment with a simple loss.

%% file: problem.tex
\section{Proposed approach: \ours}
In this work, we demonstrate the simplicity and efficiency of aligning a text encoder to a self-supervised visual foundation model. We show that we can directly use the foundation model's embedding space to perform both zero-shot classification and open-vocabulary semantic segmentation.
The first section defines the model architecture and training objective.
We then describe the data curation that establishes our training dataset, followed by our inference protocol. An overview of our method is shown in \autoref{fig:method overview}.

\subsection{Locked-image text alignment}
\noindent{\textbf{Image and text encoders.}} For the image encoder, we use a frozen ViT model \cite{dosovitskiy2020image}, denoted $\venc(\cdot)$, trained in a self-supervised fashion following DINOv2~\cite{oquab2024dinov2}. The encoder takes as input an image $\image \in \Reals^{3 \times H \times W}$
and divides it into a sequence of $N$ patch tokens to which is prepended a learned $\cls$ token, thus giving $\venc(\image) = [\mathbf{c}, \mathbf{f}_1, \cdots, \mathbf{f}_N]$ in $\Reals^{(1 + N) \times D}$. Here, $\mathbf{c} \in \Reals^D$ represents the \cls output embedding and $\mathbf{f}_p \in \Reals^D$ denotes the output embedding for a patch $p \in [1, \cdots, N]$. We discard any potential register tokens~\cite{darcet2023vision} as they are not used.

The text encoder consists of a series of Transformer~\cite{vaswani2017attention} blocks and a single linear layer on top that maps features to the image embedding size.
All parameters of the text encoder are trained from scratch following LiT~\cite{zhai2022lit}.
We follow \cite{radford2021learning, zhai2022lit} and align the output \eos text token to the image embedding, therefore obtaining global alignment between the corresponding sentences and images.

\noindent{\textbf{Improved image representation for text alignment.}} Here we describe our choice of image representation for the image-text alignment.
We aim to improve the global-level text alignment used for classification and retrieval tasks, as well as the patch-level alignment for segmentation, using \textit{a single learning objective that does not require any pixel-level supervision}.
Previous works have proposed approaches designed specifically for a task:
for image-level classification, the \cls embedding $\mathbf{c}$ is the predominant choice for text alignment~\cite{radford2021learning},
while for segmentation, the final patch embeddings can be pooled \cite{mukhoti2023pacl, clippy2022}, \eg using max or average pooling, enforcing gradient to the patches but unfortunately hurting classification performance \cite{clippy2022}.
Instead, we aim here to enforce \emph{both} global and local alignment with text.
To do so, we concatenate
the $\cls$ embedding to the average-pooled patch embeddings.

We further improve the alignment to the text modality by adding
two trainable transformer blocks, noted $\psi$, on top of the frozen vision backbone,
which we refer to as ``vision blocks'' throughout the paper.
We use the outputs of the blocks $\psi$,
which preserves the dimensionality $D$ of the descriptors,
following:
\begin{equation}
\psi([\mathbf{c},\mathbf{f_1},\cdots,\mathbf{f_N}]) = [\mathbf{c'},\mathbf{f_1'},\cdots,\mathbf{f_N'}],
\end{equation}
to produce a representation for the image. Specifically, we concatenate the updated class token $\mathbf{c'}$ with the pooled patch tokens, obtained by applying a pooling operation $\sigma$ over the patches $[\mathbf{f_1'},\cdots,\mathbf{f_N'}]$
and produce the  global descriptor $\mathbf{g}$ of dimension $\Reals^{2D}$ :
\begin{equation}
    \mathbf{g}=[\mathbf{c'}; \sigma{([\mathbf{f_1'},\cdots,\mathbf{f_N'}])}],
\end{equation}
with `;' denoting the concatenation. We found that using average pooling for $\sigma$ yields the best results.
By propagating the gradient through the average of patch tokens, each token can learn to contribute to the final descriptor, enabling a more granular alignment.
We discuss the importance of this representation for dense tasks in \autoref{tab:ablation:pool}.
Interestingly, we observe that this joint representation improves alignment for downstream classification and segmentation tasks.

\noindent{\textbf{Contrastive locked-image text alignment.}}
The image-text alignment objective encourages the text representation to be close to its paired image while simultaneously repelling it from non-corresponding images.
As discussed, we use the image descriptor $\textbf{g}$ as the alignment target to the text embedding.
With the image backbone frozen, the trainable parts consist of the additional vision blocks and the text encoder.
We train with the contrastive learning objective~\cite{radford2021learning} on a dataset of image-caption pairs, which is automatically curated without any supervision.

\subsection{Text- and image-based data curation}
\label{sec:dataset}
Training data plays a crucial role in machine learning model performance~\citep{vo_automatic_2024,xu2024demystifying}.
CLIP-style VLM training requires good quality image-text pairs, however existing data curation methods for this rely only on the text modality~\cite{radford2021learning, xu2024demystifying} which we later show is suboptimal. For example, recent MetaCLIP~\citep{xu2024demystifying} curates CLIP training data from a large pool of image-text pairs collected from the Internet.
It first constructs a set of text queries based on WordNet~\cite{miller1995wordnet} synsets and Wikipedia. Next, a mapping is establishes from each query to the set of image-text pairs whose caption contains the query. Finally, pairs are sub-sampled from each of these sets to form the final dataset.  This approach results in a more balanced distribution of concepts within the data pool.

\begin{figure}[t]
    \centering
    \includegraphics[width=\linewidth]{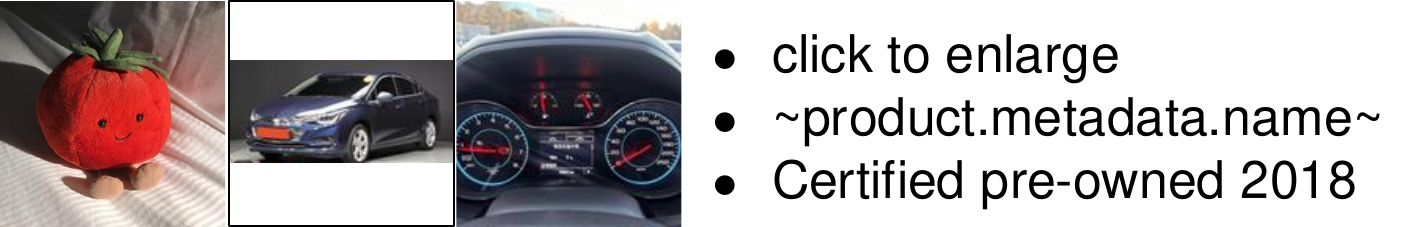}
    \vspace{-3ex}
    \caption{\textbf{Examples of poor, ambiguous or too generic captions} found in our data pool.}
    \vspace{-2ex}
    \label{fig:poor_caption}
\end{figure}

Although the pipeline significantly improves the performance of CLIP, it overlooks the distribution of visual concepts appearing in the data pool. This would not be an issue if there was a perfect alignment of concepts between the images and their captions because selecting data based on text would lead to a balanced selection in the visual space.
However, captions automatically collected from the Internet are often noisy and do not exactly describe what is depicted in the corresponding images (see~\autoref{fig:poor_caption}), therefore, ignoring the image distribution is suboptimal.

In this work, we propose to balance both the caption and image distributions by combining text and image based curation. We use \cite{xu2024demystifying} for text curation while for image curation, we use the clustering-based method of \citet{vo_automatic_2024}.
The latter divides the data pool into coherent clusters from which data are sampled
to form a curated dataset. As opposed to~\citep{xu2024demystifying}, this method
does not require a pre-defined set of concepts to perform clustering, which is not trivial to construct in the visual space. Instead it builds clusters using  hierarchical $k$-means. Clusters obtained this way
distribute evenly over the space and their size naturally follow a long-tailed distribution. The curated dataset is then formed by sub-sampling from the clusters to diminish the impact of head clusters and thus balance the concepts. In this work, we propose applying this curation method on images and the pipeline of~\cite{xu2024demystifying} on captions. We then take the intersection of these results to form the final selection.

\subsection{Inference}
\label{sec:inference}
At inference, we consider a set of text queries $\mathcal{Q}$ which we want to compare either with the image representation (\eg for classification and retrieval tasks) or to the image pixels for dense tasks. In both cases, each text query is encoded by the trained text encoder as $T_q = \tenc(q) \in \Reals^{2D}$ for $q \in \mathcal{Q}$.

In order to perform open-vocabulary zero-shot classification and retrieval, we extract a global image descriptor $\textbf{g}$ which we compare with each of the text queries using cosine similarity.
In the case of a dense task which requires pixel-level features, with our model, there is no need to adapt the output specifically to the task, \eg as done in MaskCLIP \cite{zhou2022maskclip}.
Instead, we extract for each patch
$p \in [1, \cdots, N]$
the final representation $\mathbf{f'_p}$ in $\Reals^{D}$ outputted by the model.
We then compare, using cosine similarity, each patch representation with the part of the text embedding aligned during training with the average patch.
We then upsample the logits to fit to the image resolution.
Doing so, we obtain good quality predictions without needing any model adaptation \cite{zhou2022maskclip}.
This is possible because the patch space benefits both from the SSL localization quality and the alignment to the text learned with our objective.

%% file: experiments.tex
\section{Experiments}
\subsection{Tasks and metrics}
\label{sec:tasksandmetrics}

\paragraph{Zero-shot classification.}
We evaluate zero-shot classification using the protocol described in CLIP on ImageNet-1K~\cite{imagenet_cvpr09} (IN1K), ImageNet-v2~\cite{recht2019imagenetv2} (IN-v2), ImageNet-Adversarial~\cite{hendrycks2021natural} (IN-A), ObjectNet~\cite{barbu2019objectnet} (ObjNet), iNaturalist2021~\cite{inaturalist21} (iNat21) and Places205~\cite{places205} (PL205). At test time, we feed the class names to the text encoder to retrieve text vectors, and measure their cosine similarity with the global descriptor produced by the image encoder.

\paragraph{Image-text retrieval.}
We evaluate image-text retrieval on the standard cross-modal retrieval benchmarks: COCO2017~\cite{coco2017} and Flickr30K~\cite{plummer2015flickr30k}.
These datasets comprise pairs of images and their corresponding descriptive captions.
The task involves finding the most similar image based on a text query. We use the metric Recall@1, which equals 1 if the nearest image matches the ground-truth pair, and 0 otherwise.

\paragraph{Open-vocabulary segmentation.}
We evaluate the results of \ours\xspace on the task of open-vocabulary segmentation on the datasets ADE20K~\cite{zhou2019ade20k}(ADE), Cityscapes~\cite{cordts2016cityscapes} (City.), COCO-Stuff~\cite{caesar2018cocostuff}(Stuff), PASCAL Context~\cite{mottaghi_cvpr14} (C59), and PASCAL VOC20~\cite{pascal-voc-2012}(VOC). We employ the mIoU metric (mean intersection-over-union).
In order to generate pixel-level features, we use the inference procedure detailed in \autoref{sec:inference} and additionally  follow the sliding window protocol of TCL~\cite{cha2022tcl}.
If not stated otherwise we use the final patch token embeddings produced by our model.
However, most general-purpose image-text encoders~\cite{radford2021learning, zhai2023sigmoid, sun2023eva,sun2024eva, fang2024data} do not apply supervisory signals on the final patch embeddings leading to poor output patch quality.
Therefore, we employ the well-known MaskCLIP~\cite{zhou2022maskclip} strategy to evaluate such methods on the segmentation task.
It forwards the value embeddings in the last attention layer (bypassing the final attention) resulting in patch embeddings in the aligned space.


\subsection{Implementation details}

\noindent{\textbf{Training.}}
We implement our training framework in PyTorch~\cite{pazkepytorch}.
We follow the implementation of the CLIP loss function from the OpenCLIP library~\cite{ilharco_gabriel_2021_5143773}.
We employ the \texttt{torch.compile} feature of PyTorch for maximally efficient training on Nvidia A100 GPUs with 80 GB VRAM.
The DINOv2 vision encoder initialized from \cite{darcet2023vision} is kept frozen which saves compute and memory, allowing larger batch sizes compared to CLIP, which is important as shown in \autoref{tab:ablation:layers2}.
For numbers in \autoref{tab:ablation:visualfeatures},\ref{tab:ablation:pool} we follow the CLIP recipe and set the batch size to 32K.
In other tables we use 65K for better results.
We also observe that good results can be obtained by training for 50K iterations which corresponds to 1.6 and 3.2 billion image-text pairs seen at 32K and 65K batch size respectively.
We chose this setup as default and discuss more hyper-parameter in \autoref{sec:ablation}.

\noindent{\textbf{Training dataset.}}
We apply the text and image curation process described in \autoref{sec:dataset} to an initial data pool derived from CommonCrawl~\cite{commoncrawl}, consisting of 2.3 billion image-text pairs.
We sample 650 million pairs per-epoch using our curation strategy.
For the text-based curation part, text frequencies in the data pool are precomputed offline, and data with frequent texts are stochastically dropped following \citep{xu2024demystifying}. This process keeps 900 millions pairs per epoch.
For image-based curation, we use pre-trained DINOv2 ViT-L/14 to extract embeddings for an offline 3-level hierarchical $k$-means~\citep{vo_automatic_2024} with 20M, 800K and 80K centroids respectively on each level. We similarly drop pairs whose images appear in large clusters, resulting in 1.5 billion pairs per epoch. Our final training dataset for the given epoch consists of the text-image pairs kept in both the text- and image-based curation process, hereafter noted as \dataset\xspace which stands for Large Vision Text Dataset.

\noindent{\textbf{High-resolution inference.}}
A typical segmentation protocol, popularized by TCL~\cite{cha2022tcl}, consists of applying a sliding window strategy and aggregating the segmentation results in a single prediction map.
We extend this strategy to a
high-resolution windowing procedure
in which we sample crops of various sizes (1\%, 10\%, 100\% of the total area) in a dense sliding window manner, and add noise to the coordinates, such that the crops correspond to non-rectangular quadrilaterals.
We distort the crops into squares, extract features, then project the features back onto the dense pixel grid with interpolation, and average all contributions. We cluster features using $k$-means with $k$=32, then run the zero-shot classifier on the centroids. For our results using this procedure, each pixel is visited on average 40 times, for a total of approximately 800 crops processed by the vision model in 10 seconds on an A100 GPU. This approach showcases the features at finer scales, and improving the procedure is a direction for future work. We provide results in \autoref{tab:seg} (last row) and visualization in \autoref{fig:highres_sampling}.


\subsection{Ablation study of our method \ours}
\label{sec:ablation}
We study in this section the impact of different  components of \oursnospace. In all experiments, we train models on the text-based curated dataset obtained from \dataset\ following \citet{xu2024demystifying} unless otherwise specified.

\begin{table}[ht]
    \centering
    \begin{tabular}{@{} lll c cc @{}}
        \toprule
           & Vision & && \textit{class.} & \textit{retrieval} \\
        Model   &  backbone & Arch. && IN1K & COCO \\
        \midrule
        CLIP & scratch  & ViT-L/14 && 73.0 & \bf 38.0 \\
        \midrule
        \multirow{4}{*}{LiT} & MAE~\cite{he2021masked}     & ViT-L/14 && 52.3 & 13.6 \\
        & I-JEPA~\cite{assran2023self}  & ViT-H/14 && 67.7 & 20.1 \\
        & DINO~\cite{caron2021emerging}    & ViT-B/8  && 71.3 & 26.5 \\
        & DINOv2~\cite{oquab2024dinov2}  & ViT-L/14 && \bf 78.8 & 30.2 \\
        \bottomrule
    \end{tabular}
    \vspace{-1ex}
    \caption{
        \small \textbf{Comparison of trainable vision backbone (CLIP) and frozen pre-trained SSL backbones (LiT).}
        We produce CLIP results using exactly CLIP recipe in our setup.
        All the models are trained for 50K iterations.
       }
    \label{tab:ablation:visualfeatures}
\end{table}
\vspace{-5pt}

\noindent{\textbf{LiT with SSL is not obvious.}}
In order to train a text encoder to align with DINOv2 features, we resort to LiT.
Our preliminary LiT experiments with DINOv2 ViT-L/14 vision encoder and a pre-trained BERT-base~\cite{devlin2018bert} led us to 70.0 zero-shot accuracy on IN1K when training on CC12M~\citep{changpinyo2021cc12m}.
For comparison, the original LiT paper reports 67.6 with a ViT-L/16 vision encoder and pre-trained BERT-large text encoder when training on the same dataset.
We next train our models on the larger dataset \dataset, and
present in \autoref{tab:ablation:visualfeatures} a comparison between CLIP and LiT with different pre-trained vision encoders.
We observe that among considered vision backbones, DINOv2 leads to the best performance.
It enables LiT to achieve good results in classification. However, there is a drop in retrieval performance compared to CLIP, likely due to the frozen vision encoder not being able to adapt to new training data.
These results suggest that we need a new strategy to align a frozen backbone encoder to text that can generalize for different tasks, such as our proposed \oursnospace.

\begin{table}[h!t]
    \centering
    \begin{tabular}{@{}l c ccc@{}}
       \toprule
        & & \textit{class.} & \textit{retrieval} & \textit{seg.} \\
      $\sigma$ pooling && IN1K & COCO & ADE \\
       \midrule
        \cls            && 78.8 & 30.2 & 8.3 \\
        \texttt{[avg]}  && 74.7 & 32.7 & 13.3 \\
        \texttt{[max]}  && 70.2  & 25.7 & 18.0 \\
        \texttt{[CLS max]} && 78.2 & 31.9 & 16.8 \\
        \texttt{[CLS avg]} && \bf 79.2 & \bf 34.7 & \bf 18.2 \\
       \bottomrule
    \end{tabular}
    \vspace{-1ex}
    \caption{\textbf{Impact of the pooling operation $\sigma$} used at training in \ours on zero-shot performance. Results of the first row are produced with MaskCLIP strategy, others with the output patches. The experiment corresponds to the first row in \autoref{tab:ablation:layers2}.
    }
    \vspace{-10pt}
    \label{tab:ablation:pool}
\end{table}

\paragraph{Impact of the pooling operation $\sigma$ at training.}
We evaluate in \autoref{tab:ablation:pool} the impact of the choice of the pooling operation, applied during training, to the performance on downstream tasks. Typically, methods have used max \cite{clippy2022} or average pooling \cite{mukhoti2023pacl} in order to align patch embeddings to the text, but this hurts classification results. We observe the same phenomenon in our experiments where max or average pooling alone degrades classification performance. However, when using our proposed concatenation pooling \texttt{[CLS avg]}, we obtain a significant boost for \emph{both} classification and dense tasks, showing that there is no need to choose between one task or the other.

\begin{table}[ht!]
    \centering
    {\small
    \begin{tabular}{l cc ccc }
        \toprule
        && \textit{class.} & \textit{retrieval} & \textit{seg.} \\
        Ablation  && IN1K & COCO & ADE \\
        \midrule
        Reference           && 78.8 & 30.2 & 8.3 \\
        \midrule
        + avg-\pool          && 79.2 & 34.7 & 18.2 \\
        + 65K batch size     && 79.8 & 35.1 & 18.2 \\
        + 1 vision block     && 79.8 & 40.8 & 20.5 \\
        + 2 vision blocks    &&  79.7 & 42.1 & 20.4 \\
        + text-large ($768 \rightarrow 1280$) && 80.8 & 43.9 & 20.5 \\
        + img-based curation      && \bf 81.4 & \bf 45.4 & \bf 20.6 \\
        \bottomrule
    \end{tabular}
    }
    \vspace{-1ex}
    \caption{
        \textbf{Exploration of the \ours recipe.}
        We start from a `Reference' configuration which consists of training LiT with a frozen DINOv2-ViT-L/14 vision encoder and a BERT-base sized text encoder trained from scratch; we then add modifications progressively.
        The first row is evaluated following MaskCLIP whilst the next ones use the output patch tokens.
    }
    \label{tab:ablation:layers2}
\end{table}

\vspace{-25pt}
\paragraph{Improved training recipe.}
We evaluate the impact of the different components of our training in \autoref{tab:ablation:layers2}.
Beside our \pool\xspace strategy,
we observe that using a larger batch size also improves results on all tasks.
Interestingly, the addition of two learnable vision blocks on top of the vision encoder significantly improves retrieval results, showing that the task requires a better visual alignment to the text.
Increasing the text embedding size from $768$ to $1280$ also induces a large gain on all tasks.
Finally, we can observe the importance of the combined image- and text-based data curation, which is further discussed below.

\begin{table}[ht!]
    \centering
    {\small
    \begin{tabular}{@{} l cc c ccc @{}}
    \toprule
        & \multicolumn{2}{c}{curation} && \textit{class.} & \textit{retrieval} & \textit{seg.} \\
        & image  & text && IN1K & COCO & ADE \\
        \toprule
        & & && 80.3 & 42.9 & 20.0 \\
        & & \checkmark && 80.8 & 43.9 & 20.5 \\
        & \checkmark & && 80.9 & 43.7 & 20.4 \\
        & \checkmark & \checkmark && \bf 81.4 & \bf 45.4 & \bf 20.6 \\
        \bottomrule
    \end{tabular}
    }
    \vspace{-2ex}
    \caption{\textbf{Impact of the image- and text-based curation}.}
    \vspace{-25pt}
    \label{tab:ablation:curation}
\end{table}

\begin{table*}[ht!]
    \small
    \centering
    \begin{tabular}{@{} lll c cccccc c c c @{}}
        \toprule
        \multicolumn{3}{c}{} && \multicolumn{6}{c}{\it classification} && \multicolumn{2}{c}{\it retrieval} \\
        \cmidrule{5-10} \cmidrule{12-13}
        Method  & Res. & Dataset && IN1K & IN-v2 & IN-A & ObjNet & iNat21 & PL205  && COCO & Flickr \\
        \midrule
        OpenCLIP~\cite{ilharco_gabriel_2021_5143773}    & 224   & LAION-2B      && 74.0 & 66.4 & 48.0 & -- & -- & -- && 46.1 & 75.0 \\
        CLIP~\cite{radford2021learning}        & 336   & WIT-400M      && 76.6 & 70.9 & 77.5 & 72.3 & 5.7 & 59.2 && 37.1 & 67.3  \\
        MetaCLIP~\cite{xu2024demystifying}    & 224   & MetaCLIP-2.5B && 79.2 & 72.6 & 72.3 & 75.3 &10.2 & 61.8 && 45.5 & 76.9 \\
        EVA-02-CLIP~\cite{sun2023eva} & 336   & Merged-2B     && 80.4 & 73.8 & 82.9 & \bf 78.5 & 6.4 & 61.7 && 47.9 & 77.9 \\
        DFN~\cite{fang2024data}         & 224   & DFN-2B        && 81.4 & 74.5 & 66.8 & 74.1 & 17.6 & \bf 62.0 && 47.0 & 76.0 \\
        SigLIP~\cite{zhai2023sigmoid}      & 256   & WebLi         && 80.4 & 73.8 & 62.0   & 66.6$^\dagger$ & 14.8 & 58.6 && 51.1 &79.9 \\
        SigLIP~\cite{zhai2023sigmoid}      & 384   & WebLi         && \bf 82.1 & 75.9 & 76.4   & 74.0$^\dagger$ & 17.5 & 59.6 && \bf 52.7 & \bf 81.9\\
        \rowcolor{Yellow!40!white} \ours       & 224   & \dataset  && 81.4 & 75.7 & 80.0 & 72.4 & 18.8 & 61.2  && 45.4 & 77.1 \\
        \rowcolor{Yellow!40!white} \ours       & 336   & \dataset  && 81.6 & \bf 75.9 & \bf 83.2 & 74.5 & \bf 19.4 & 61.2 && 44.9 & 77.6 \\
        \bottomrule
    \end{tabular}
    \vspace{-1ex}
    \caption{
        \textbf{Zero-shot classification and retrieval} results with ViT-L models.
         $^\dagger$ indicates that the SigLIP results on ObjectNet
        (authors report 77.9 and 81.0) could not be reproduced despite obtaining matching results on ImageNet-1K.
        }
    \label{tab:cls_ret}
    \vspace{-2ex}
\end{table*}

\begin{table}[th!]
    \centering
    \resizebox{\linewidth}{!}{
    {\small
    \addtolength{\tabcolsep}{-0.4em}
    \begin{tabular}{@{}llll c ccccc@{}}
        \toprule
        Method  & Base & Train data. && ADE & City. & VOC & Stuff & C59 \\
        \midrule
         \rowcolor{gray!20!white} \multicolumn{9}{l}{\textit{Trained models specialized for segmentation}} \\
          \rowcolor{gray!20!white} GViT~\cite{xu2022groupvit}& S/16 & CC12M+RC && 9.2 & 11.1 & 79.7 & 15.3 & 23.4  \\
      \rowcolor{gray!20!white} CoCu~\cite{NEURIPS2023_d937cb3f} & S/16 & CC12+3M+Y & & 12.3& 22.1 & 51.4 & 15.2 & 23.6  \\
        \rowcolor{gray!20!white} CLIPpy~\cite{clippy2022} & B/16 & H-134M & & 13.5& - & 52.2 & - &-  \\
        \rowcolor{gray!20!white} TCL~\cite{cha2022tcl} & B/16 & CC12+3M &&  14.9 & 23.1 & 77.5 & 19.6 & 30.3 \\
        \rowcolor{gray!20!white} CoDe~\cite{wu2024imagetextcodecompositiontextsupervisedsemantic} & B/16 & CC12+3M & & 17.7 & 28.9 & 57.7 & 23.9 & 30.5  \\
        \midrule
        \multicolumn{7}{l}{\textit{Generalist image-text encoders}} \\
        MetaCLIP~\cite{xu2024demystifying}    & L/14   & MCLIP-2.5B && 2.5 & 1.7 & 23.3 & 3.9 & 6.5 \\
        CLIP~\cite{radford2021learning}    & B/32   & WIT-400M          &&  5.0 & 8.6 & 34.7 & 9.0 & 14.2 \\
        DFN~\cite{fang2024data}         & L/14   & DFN-2B        && 5.8 & 8.0 & 25.9 & 5.1 & 10.0 \\
        OpenCLIP~\cite{ilharco_gabriel_2021_5143773}    & L/16   & LAION-2B      && 5.9 & 9.8 & 30.0 & 8.3 & 13.1 \\
        CLIP~\cite{radford2021learning}    & L/14   & WIT-400M &&  6.0 & 11.5 & 24.8 & 7.3 & 10.9 \\
        SigLIP~\cite{zhai2023sigmoid} & L/14   &  WebLi  && 9.1 & 18.3 & 30.3 & 9.5 & 13.7 \\
        OpenCLIP~\cite{ilharco_gabriel_2021_5143773}    & B/32  & LAION-2B      && 9.9 & 18.1 & 42.9 & 12.7 & 19.0 \\
        OpenCLIP~\cite{ilharco_gabriel_2021_5143773}    & B/16  & LAION-2B      && 12.7 & 20.2 & 45.4 & 16.4 & 24.2 \\
        \rowcolor{Yellow!40!white} \ours                             & L/14 & \dataset && \bf 20.6 & \bf 32.1 & \bf 62.1 & \bf 20.9 & \bf 30.9 \\
       \rowcolor{Yellow!40!white}  HR(\oursnospace) & L/14 & \dataset && \textit{25.1} & \textit{41.0} & \textit{67.6} & \textit{24.1} &  \textit{36.7} \\
        \bottomrule
    \end{tabular}
    }
    }
    \caption{
        \textbf{Open-vocabulary segmentation} performance in mIoU (\%). All `generalist image-text encoders' methods are evaluated using MaskCLIP defined in \autoref{sec:tasksandmetrics}.
        We put in gray methods specialized for segmentation and bold the sections separately.
        For reference, we also produce results with our
        high-resolution inference procedure (noted `HR' and in italic).
    }
    \label{tab:seg}
    \vspace{-10pt}
\end{table}

\paragraph{Impact of dataset curation.}
\autoref{tab:ablation:curation} decouples the impact of each data curation strategy on \ours results.
Both text- and image-based data curation help to re-balance the long-tailed data distribution, and boost performance.
Our proposed combination of them leads to the best performance on all of the three tasks. This result highlights the important of curating data based on both text and visual modality for visual-language training.


\subsection{Comparisons to state of the art}
\label{sec:zero-shot-classification-retrieval-segmentation}
\paragraph{Zero-shot classification and retrieval.}
We compare \ours with state-of-the-art baselines in~\autoref{tab:cls_ret} on two image-level understanding tasks: zero-shot image classification and cross-modal retrieval.
Our model is on par or better than alternative CLIP-like models on classification benchmarks, setting the state-of-the-art performance on IN-v2, IN-A and the challenging iNaturalist datasets. It can also be observed that the performance of \ours is lower than competitors such as SigLIP~\citep{zhai2023sigmoid} on text-image retrieval tasks. This is likely due to the unsatisfactory quality of our trained text encoder, which in turn is a consequence of freezing the vision encoder, as discussed later in \autoref{subsec:perf-text-tower}. However, we see next that \ours largely outperforms SigLIP in open-vocabulary segmentation task.


\begin{figure}[ht!]
    \centering
\setlength{\tabcolsep}{1pt}
    \begin{tabular}{ccc}
        \includegraphics[height=0.32\linewidth]{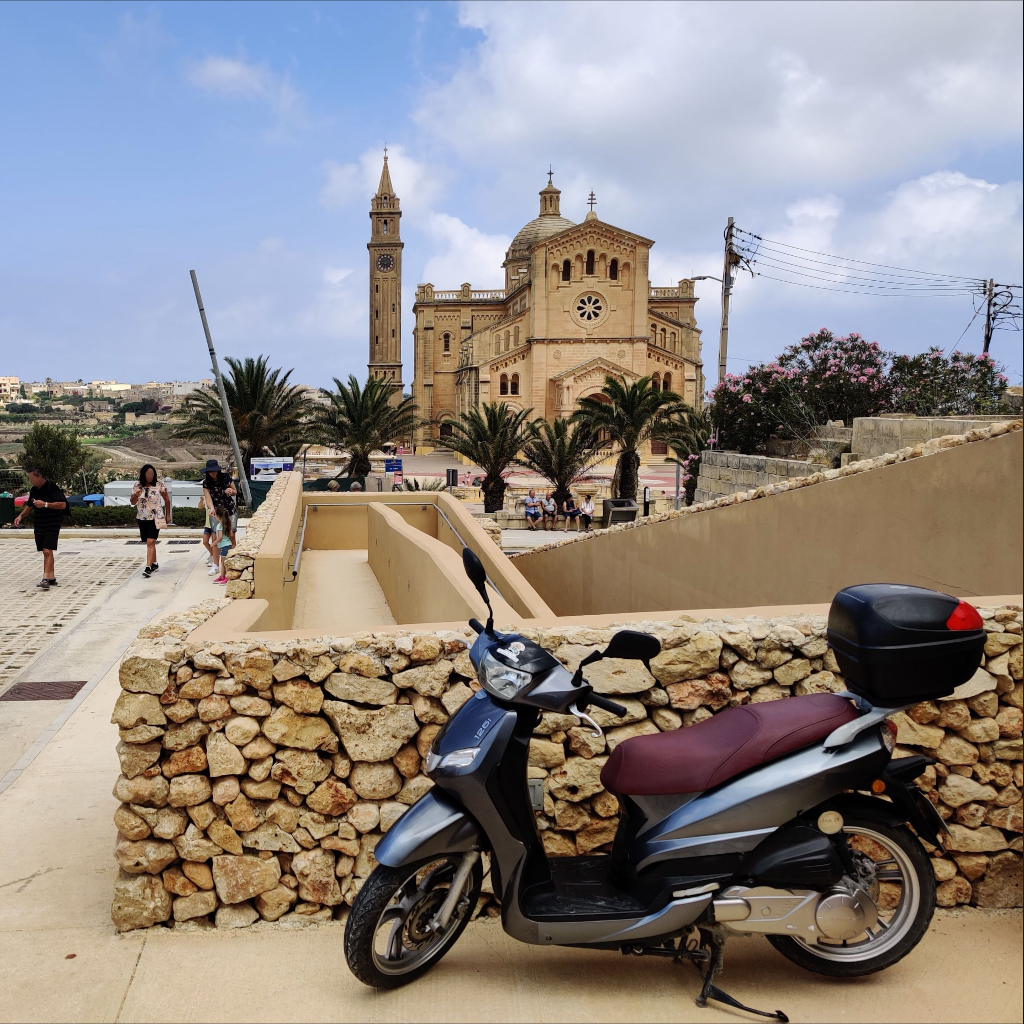} &
        \includegraphics[height=0.32\linewidth]{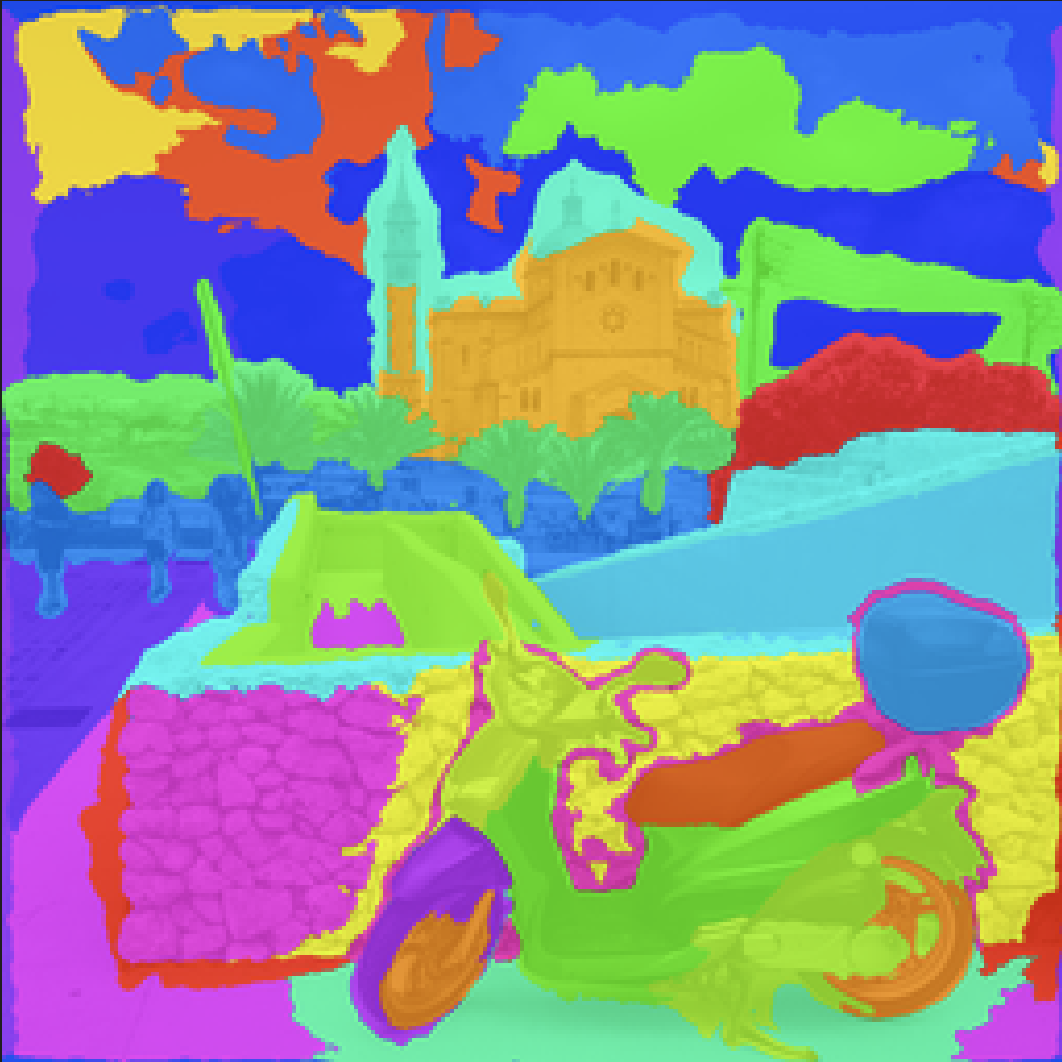} &
        \includegraphics[height=0.32\linewidth]{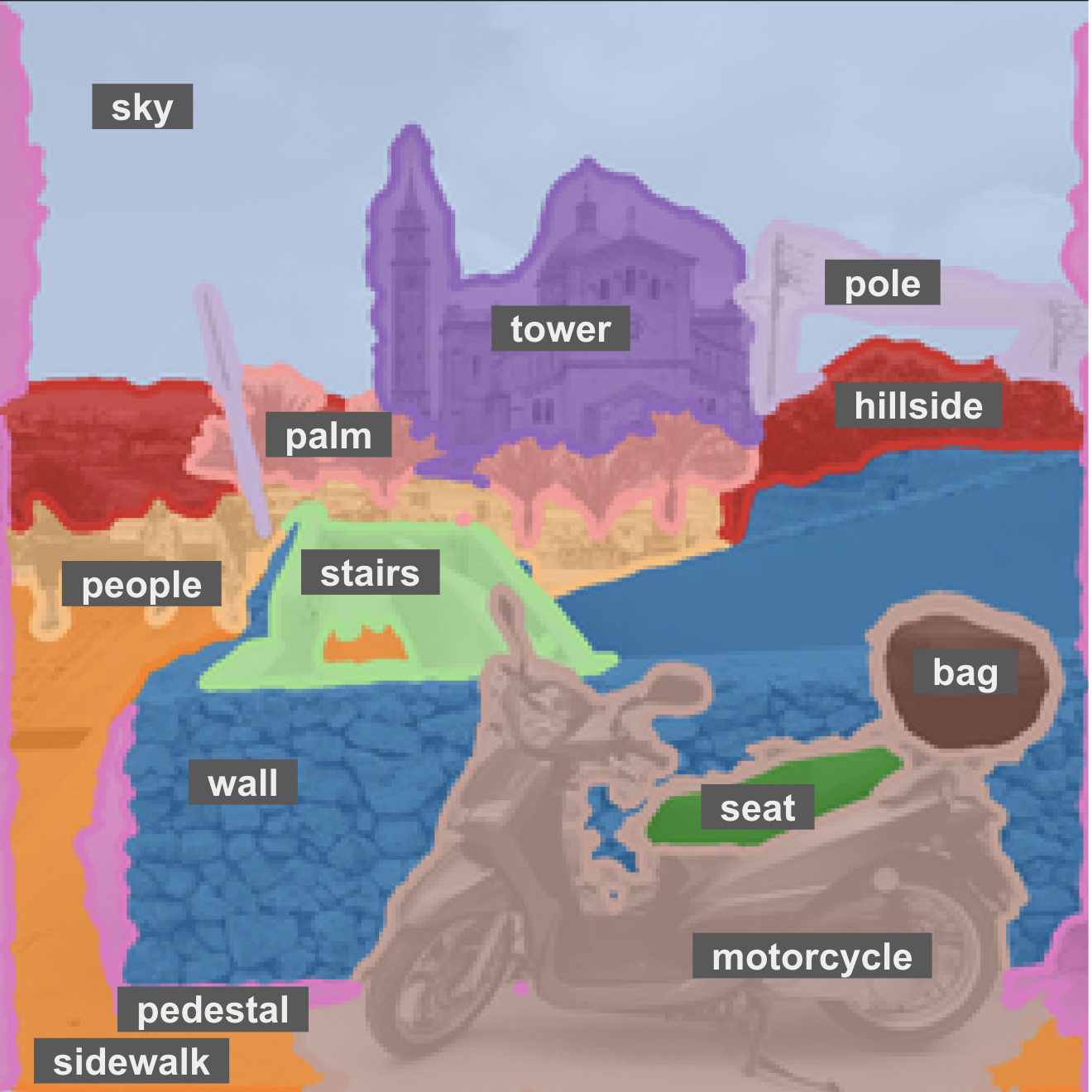} \\
        \small Image & \small $k$-means & \small Predictions
    \end{tabular}
   \vspace{-5pt}

    \caption{\textbf{High-resolution inference.} \textbf{Left}: input image. \textbf{Middle}: result of $k$-means clustering ($k$=32) on the features. \textbf{Right}: open-vocabulary predictions with the ADE20K class names.}
    \label{fig:highres_sampling}
    \vspace{-20pt}
\end{figure}

\paragraph{Open-vocabulary segmentation.}
On segmentation, our approach greatly outperform alternatives as shown in \autoref{tab:seg}, and performs on par or better than specialized models, without any engineering refinement: we simply apply the classification model on the local features. We note that, compared to other methods, the performance trends lower on VOC20 while being higher on other datasets, which we attribute to a domain gap, as VOC20 frequently contains only one close-up centered object.
We ablate in the appendix the impact of using our proposed representation versus MaskCLIP. We also show the quality of our results with our high-resolution strategy in \autoref{fig:highres_sampling}.

\paragraph{Training efficiency.}
We show in \autoref{fig:intro-fig} how zero-shot classification performance on the IN1K validation set evolves as a function of training GFLOP for \ours and CLIP, trained on the same \dataset\xspace dataset, described below. On 128 A100 GPUs, 19 hours of training are enough to reach 81.4\% on IN1K. In comparison, CLIP requires 110 hours to obtain 79.0\%. Furthermore, restricting CLIP training to match the GFLOP budget of our best performing model only achieves 73\% accuracy, 8.4\% below our model.

%% file: error_analysis.tex
\subsection{Further Analysis}
\label{sec:error_analysis}

\paragraph{Failure modes.} We conduct an error analysis on the open-vocabulary segmentation task, on the ADE20K dataset.

\noindent{\textit{Object boundaries.}} We replace the $k$-means operation in our inference procedure by the ground-truth masks to perform open-vocabulary predictions. This leads to a perfect-boundary topline of 38.9 mIoU on ADE20K, meaning that the remaining performance gap can be attributed to the misalignment between image and text. In this case, errors include predicting ``shower'' where the real label is ``wall'', in a bathroom photo, suggesting the patch features take the context into account to some extent.

\noindent{\textit{Object overlaps.}}
We observed multiple cases where overlapping objects are predicted but not annotated due to overlaps. For example, in \autoref{fig:highres_sampling} we can observe that the seat of the motorbike is predicted separately, while in usual annotations the whole motorbike is labeled as a single entity. This decreases the benchmark score for this class of models, solely due to the dataset collection procedure.

\noindent{\textit{Class names.}} We observe during evaluation that some class names do not always correspond best to their meaning in everyday language as found in captions for image-text paired data. For example, the building class of ADE20K almost systematically corresponds to a facade; the class name ``person'' is rarely used in captions, while ``people'' is more frequent. Similarly, the ``vegetation'' class name in CityScapes is problematic. To account for these discrepancies, we search for the optimal class names on ADE20K by averaging the token embeddings for the ground-truth masks, and using the closest word in the embedding space. We obtain a new list of class names, that we present in appendix.
This procedure can add +2.1 mIoU on ADE20K
which shows that the dataset class names, chosen arbitrarily at collection time, have a strong impact on the result.

This suggests that existing datasets are not well-suited to evaluate open-vocabulary semantic segmentation: first because classes naturally overlap (windows are often included in buildings), second because the class names are not aligned with their use in common natural language.

\paragraph{Quality of the text encoder.}
\label{subsec:perf-text-tower}
We analyze the quality of our trained text encoder by evaluating it on text classification, clustering, reranking, and pair classification tasks in the text embedding benchmark suite MTEB~\citep{muennighoff2022mteb}.
Our text encoder is outperformed by CLIP's text encoder by a margin of 4.2\% on average on these tasks.
Moreover, when comparing \ours with and without two learnable blocks on top of the vision encoder, we observe that removing the blocks further decreases the performance by 3.2\% on average.
These results show that freezing the vision encoder might hurt the text encoder and lead to lower performance on tasks such as image-text retrieval.
They also suggest that we need to find a better trade-off between exploiting the quality of frozen vision encoder and allowing it to adapt to new data domain, left for future work.

%% file: supp.tex
\section{Additional details}

\paragraph{Computational cost.}
In this work, we show that we can obtain good global and local vision-language alignment with minimal additional cost thanks to powerful pre-trained SSL models.
This appears to be a more efficient paradigm than training CLIP from scratch. The computational costs for training our models and different CLIP models are reported in \autoref{tab:computation}.
For completeness' sake, we also include the pretraining cost of the ViT-g DINOv2 vision encoder as well as the cost of distilling this model into a ViT-L. In practice, such additional costs should however be considered amortized over the multiple downstream adaptations of the DINOv2 backbone.

\begin{table}[ht!]
    \centering
    \resizebox{\linewidth}{!}{
    \begin{tabular}{l c cccccc}
        \toprule
         && Samples  & Batch  & GPUs & total & GPU \\
        Method &&  seen &  size &  &  GPU.h & arch. \\
        \midrule
        CLIP && 12.8B & 32768 & 256 & 73728 & V100  &\\
        OpenCLIP && 12.8B & 38400 & 400 & 50800 & A100 40 GB  &\\
        MetaCLIP && 12.8B & 32768 & 128 & 92160 & V100  &\\
        EVA-02-CLIP && 2B & 61000 & 128 & -- & A100 40 GB  &\\

        \midrule
        DINOv2 ViT-g pretraining &&  -- & -- & 256 & 22000 & A100 80 GB\\
        DINOv2 ViT-L distillation && -- &  -- & -- & 8000 & A100 80 GB\\
        \ours && 3.2B & 65536 & 128 & 2432 & A100 80 GB\\
        \ours@336 && 3.2B & 65536 & 256 & 4096 & A100 80 GB\\
        \bottomrule
    \end{tabular}
    }
    \caption{\textbf{Computational cost of different models in GPU hours.}}
    \label{tab:computation}
\end{table}

\paragraph{ADE20K class names for the error analysis discussion.}

In \autoref{sec:error_analysis}, we discuss the failure modes of our zero-shot semantic segmentation method.
In particular, we show that class names can be optimized to boost results, instead of using the default ones from each dataset.
This is not surprising, the 150 class names of ADE20K were originally chosen to identify each category and were not intended as holistic descriptors for zero-shot segmentation via a vision-language model.
In our experiments, we have observed that some class names are too broad, \eg \textit{building}, or ambiguous, \eg \textit{throne}, and consequently result in incorrect predictions.
In \autoref{supmat:classnames}, we include the optimized class names for ADE20K that improve open-vocabulary segmentation by 2.1 mIoU points, as reported in the discussion about failure modes in \autoref{sec:error_analysis}.
Please note that for all experiments in the main text, we use the original class names to facilitate comparison with previous work.

\section{Additional ablation studies}

\begin{table}[!h]
    \centering
    \small{
    \begin{tabular}{lc c cc}
       Inference && \multicolumn{2}{c}{\textit{segmentation}} \\
        embedding  && ADE & City. \\
        \toprule
        \texttt{[value]} (MCLIP)     &&   7.0 & 11.7 \\
        \texttt{[CLS patch]}          && 19.9  & 26.2 \\
        \texttt{[value patch]}       &&  20.0 &  29.0 \\
        \texttt{[patch]}              && \bf 20.6 & \bf 32.1 \\
        \bottomrule
    \end{tabular}
    }
    \caption{
        \textbf{Ablation of the embedding in dense zero-shot segmentation inference.} We show segmentation results with different embeddings to represent a patch, on the datasets ADE20K and Cityscapes. `MCLIP' corresponds to MaskCLIP~\cite{zhou2022maskclip} strategy, which we also name here \texttt{value}.
    }
    \label{tab:ablation:seg}
\end{table}

\paragraph{Impact of the embedding in segmentation.} \autoref{tab:ablation:seg} presents open-vocabulary segmentation results on the challenging datasets ADE20K and Cityscapes. We follow the evaluation protocol of TCL~\cite{cha2022tcl}.
Following only MaskCLIP patch representation (\texttt{[value]}) leads to the worst results.
Using solely the model's output patch descriptor (\texttt{[patch]}) and their corresponding part in the text embedding leads to the best results. This is the setup used in the main paper. We also observe that concatenating the \texttt{[CLS]} token to the patch representation hurts the performance \vs \texttt{[patch]} only, particularly in Cityscapes: we found this to be due to the dominance of the salient visual concept in the \texttt{[CLS]}.

\paragraph{Impact of the image embedding size at training.} We show in \autoref{tab:embed_size_ablation} that the benefit of using the concatenated representation $\mathbf{g}$ (noted here \texttt{[CLS avg]}) when training \ours does not come from higher dimensionality of the image embedding. To this end, we have conducted an additional experiment in which we project the \texttt{[CLS]} token from the dimension of 1024 to 2048 before passing it to the vision blocks.
Little impact is observed from this dimensionality change. This additionally shows that the gain (from 30.9 to 34.7) in the retrieval task is largely due to the concatenation of the \cls token with \texttt{[avg]}.

\begin{table}[ht!]
    \centering
    \small{
    \begin{tabular}{lc c cc}
      Training &  && \textit{class.} & \textit{retr.} \\
     embedding & proj && IN1K & COCO \\
    \toprule
        \cls & && 78.8 & 30.2 \\
        \cls & $1024 \rightarrow 2048$ & &78.8 & 30.9 \\
         \texttt{[CLS avg]} & && \bf 79.2 & \bf 34.7  \\
    \bottomrule
    \end{tabular}
    }
    \caption{\textbf{Analysis of the image embedding size at training time.} Projecting the \cls embedding to dimension 2048 (second row) yields minimal gain on the benchmarks.
    }
    \label{tab:embed_size_ablation}
\end{table}

\section{Additional qualitative results}

\begin{figure*}
  \begin{minipage}[t]{.34\linewidth}
    \centering
    \includegraphics[width=1\linewidth]{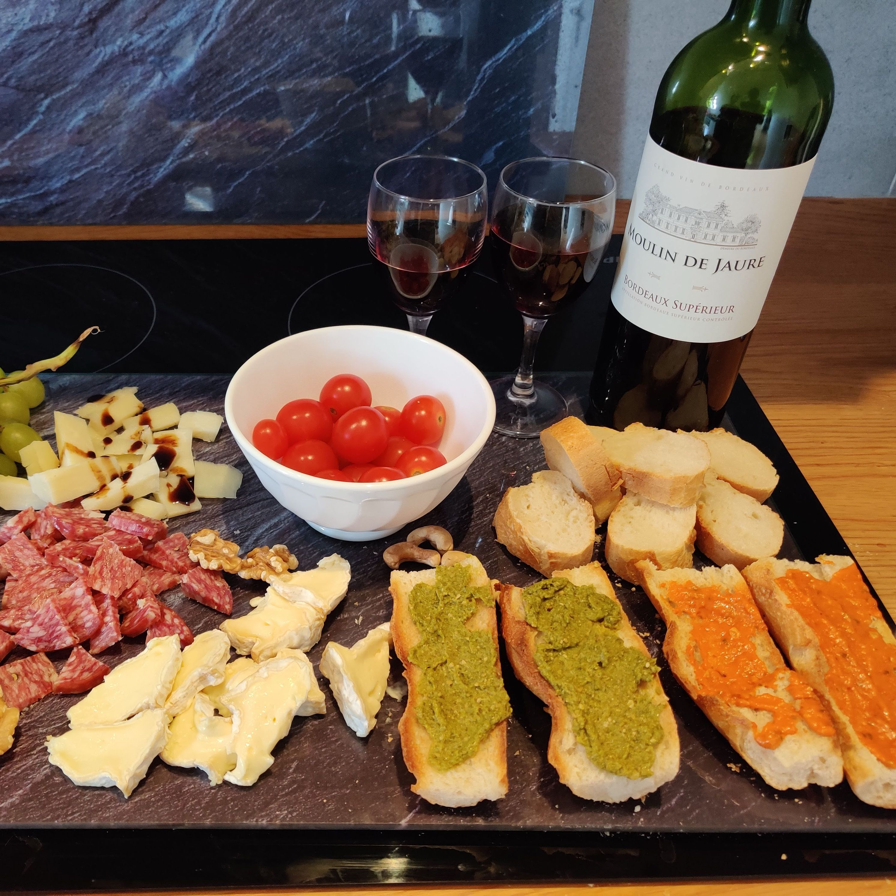}
  \end{minipage}\hfill
  \begin{minipage}[t]{.34\linewidth}
    \centering
    \includegraphics[width=1\linewidth]{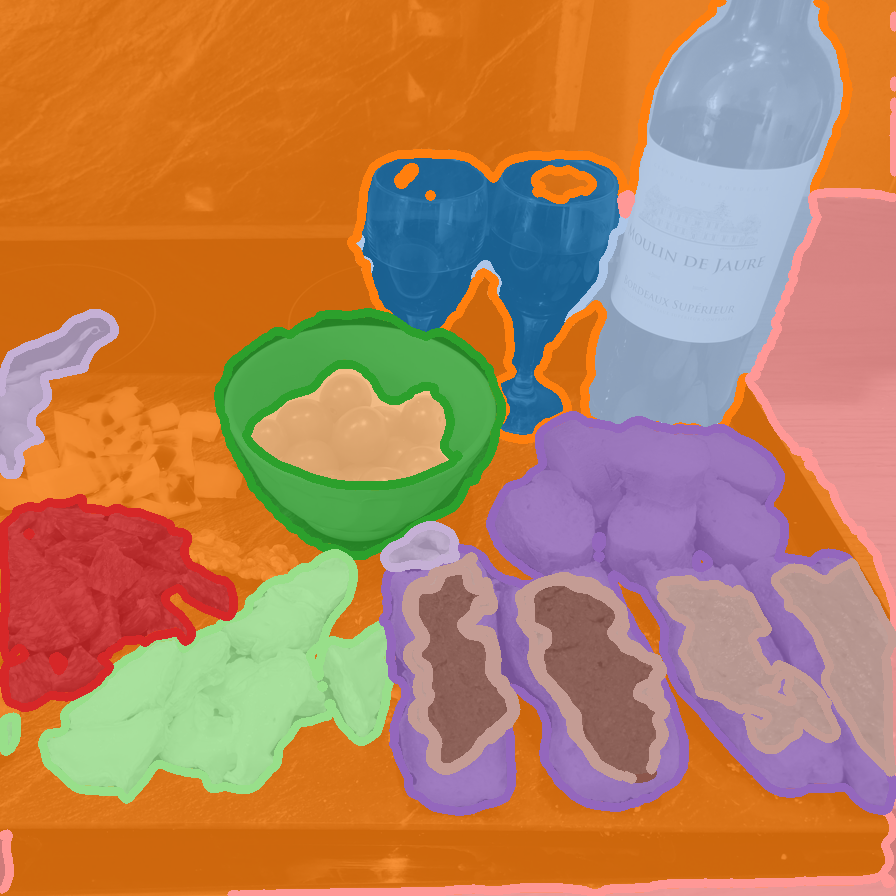}
  \end{minipage}\hfill
  \begin{minipage}[t]{.30\linewidth}
    \centering
    \small
    \begin{tabular}[b]{lll}
        \toprule
        & Color & Name \\
        \midrule
        & \cellcolor[HTML]{1f77b4} & Wine glass \\
        & \cellcolor[HTML]{aec7e8} & Wine bottle \\
        & \cellcolor[HTML]{ff7f0e} & Stone cutting board \\
        & \cellcolor[HTML]{ffbb78} & Cherry tomatoes \\
        & \cellcolor[HTML]{2ca02c} & White ceramic bowl \\
        & \cellcolor[HTML]{98df8a} & French cheese \\
        & \cellcolor[HTML]{d62728} & Salami slices \\
        & \cellcolor[HTML]{ff9896} & Wooden table \\
        & \cellcolor[HTML]{9467bd} & Sliced baguette \\
        & \cellcolor[HTML]{c5b0d5} & Green grapes \\
        & \cellcolor[HTML]{8c564b} & Pesto bruschetta \\
        & \cellcolor[HTML]{c49c94} & Red pepper spread on bread \\
        \bottomrule
    \end{tabular}
  \end{minipage}
  \begin{minipage}[t]{.34\linewidth}
    \centering
    \includegraphics[width=1\linewidth]{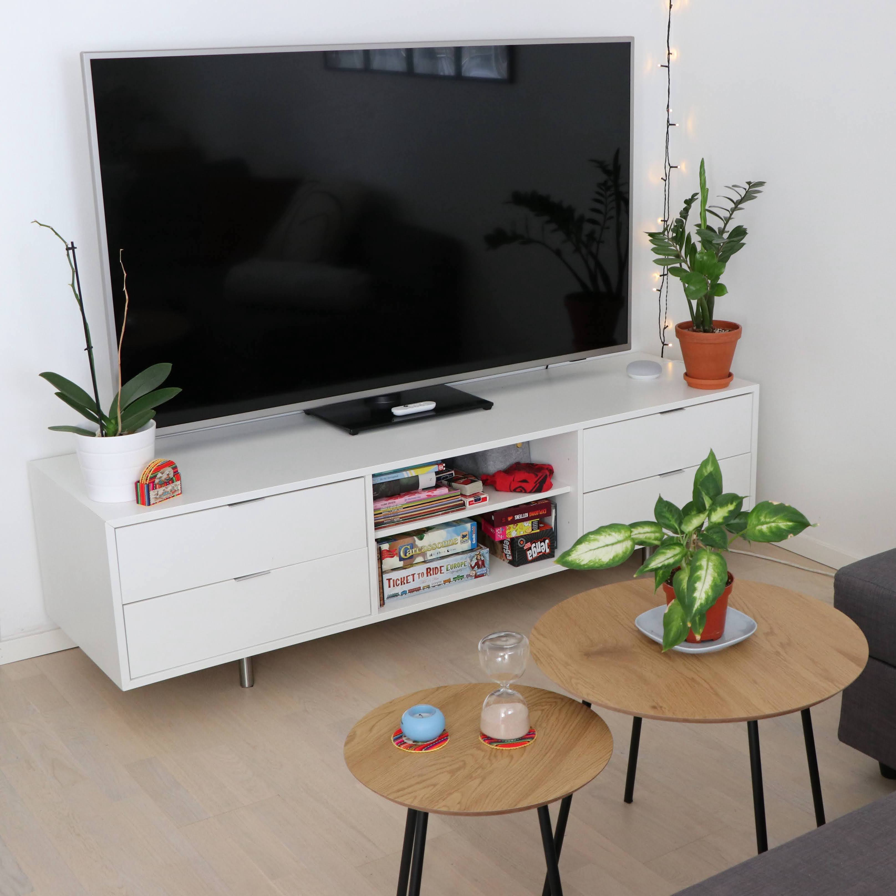}
  \end{minipage}\hfill
  \begin{minipage}[t]{.34\linewidth}
    \centering
    \includegraphics[width=1\linewidth]{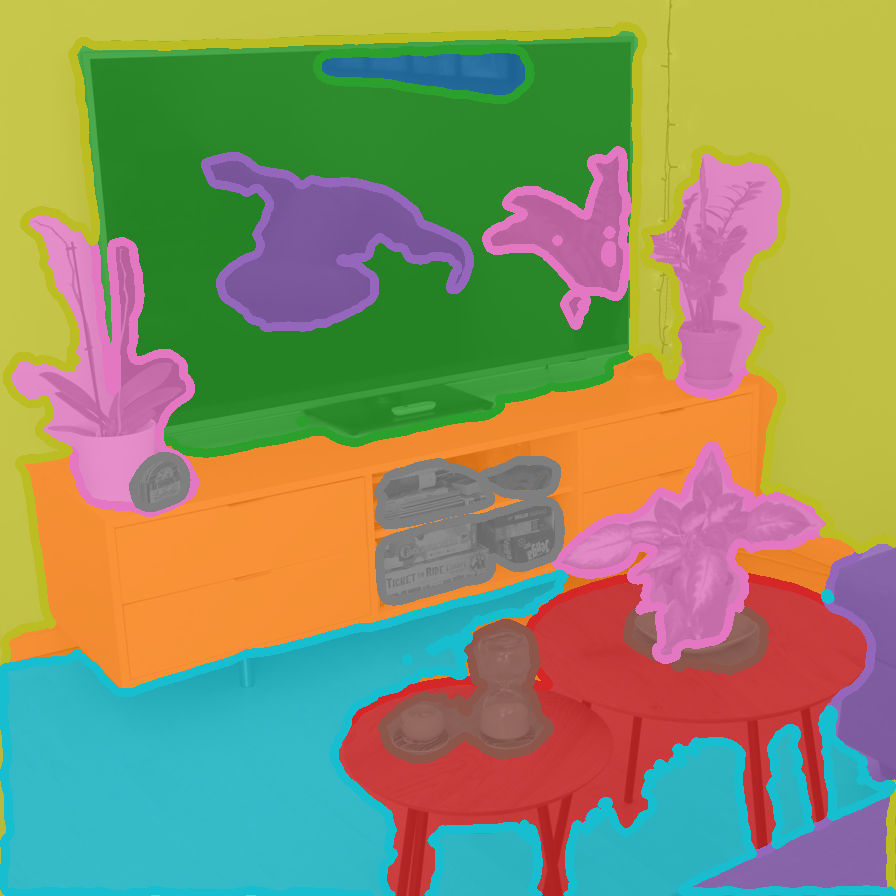}
  \end{minipage}\hfill
  \begin{minipage}[t]{.30\linewidth}
    \centering
    \small
    \begin{tabular}[b]{lll}
        \toprule
        & Color & Name \\
        \midrule
        & \cellcolor[HTML]{1f77b4} & Window \\
        & \cellcolor[HTML]{ff7f0e} & White cabinet \\
        & \cellcolor[HTML]{2ca02c} & Black television screen \\
        & \cellcolor[HTML]{d62728} & Wooden sofa table \\
        & \cellcolor[HTML]{9467bd} & Gray couch \\
        & \cellcolor[HTML]{8c564b} & Candle \\
        & \cellcolor[HTML]{e377c2} & Potted plant \\
        & \cellcolor[HTML]{7f7f7f} & Books \\
        & \cellcolor[HTML]{bcbd22} & Indoor wall \\
        & \cellcolor[HTML]{17becf} & Parquet floor \\
        \bottomrule
    \end{tabular}
  \end{minipage}
  \begin{minipage}[t]{.34\linewidth}
    \centering
    \includegraphics[width=1\linewidth]{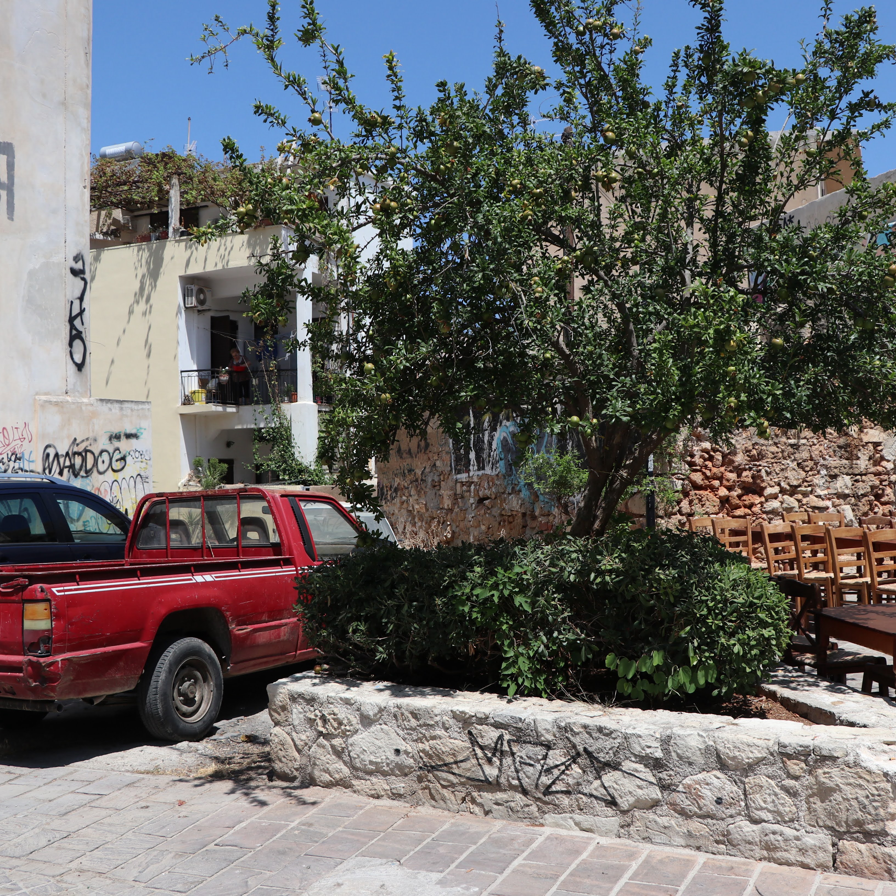}
  \end{minipage}\hfill
  \begin{minipage}[t]{.34\linewidth}
    \centering
    \includegraphics[width=1\linewidth]{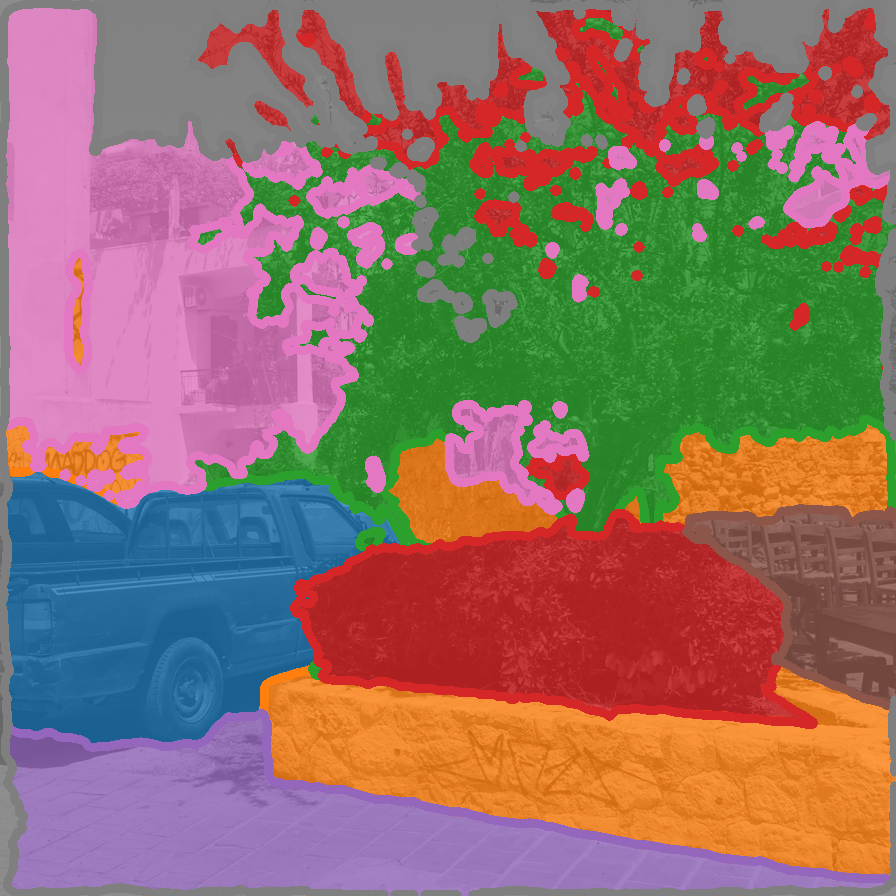}
  \end{minipage}\hfill
  \begin{minipage}[t]{.30\linewidth}
    \centering
    \small
    \begin{tabular}[b]{lll}
        \toprule
        & Color & Name \\
        \midrule
        & \cellcolor[HTML]{1f77b4} & Red pickup truck \\
        & \cellcolor[HTML]{ff7f0e} & Stone wall \\
        & \cellcolor[HTML]{2ca02c} & Lush tree \\
        & \cellcolor[HTML]{d62728} & Bush \\
        & \cellcolor[HTML]{9467bd} & Paved road \\
        & \cellcolor[HTML]{8c564b} & Chair \\
        & \cellcolor[HTML]{e377c2} & Facade \\
        & \cellcolor[HTML]{7f7f7f} & Blue sky \\
        \bottomrule
    \end{tabular}
  \end{minipage}
  \caption{\textbf{Open-vocabulary semantic segmentation, part 1/2.} The input resolution is $896{\times}896$ pixels.}
  \label{fig:supmat-open-vocab-segmentation-part-2}
\end{figure*}

\begin{figure*}
  \begin{minipage}[t]{.34\linewidth}
    \centering
    \includegraphics[width=1\linewidth]{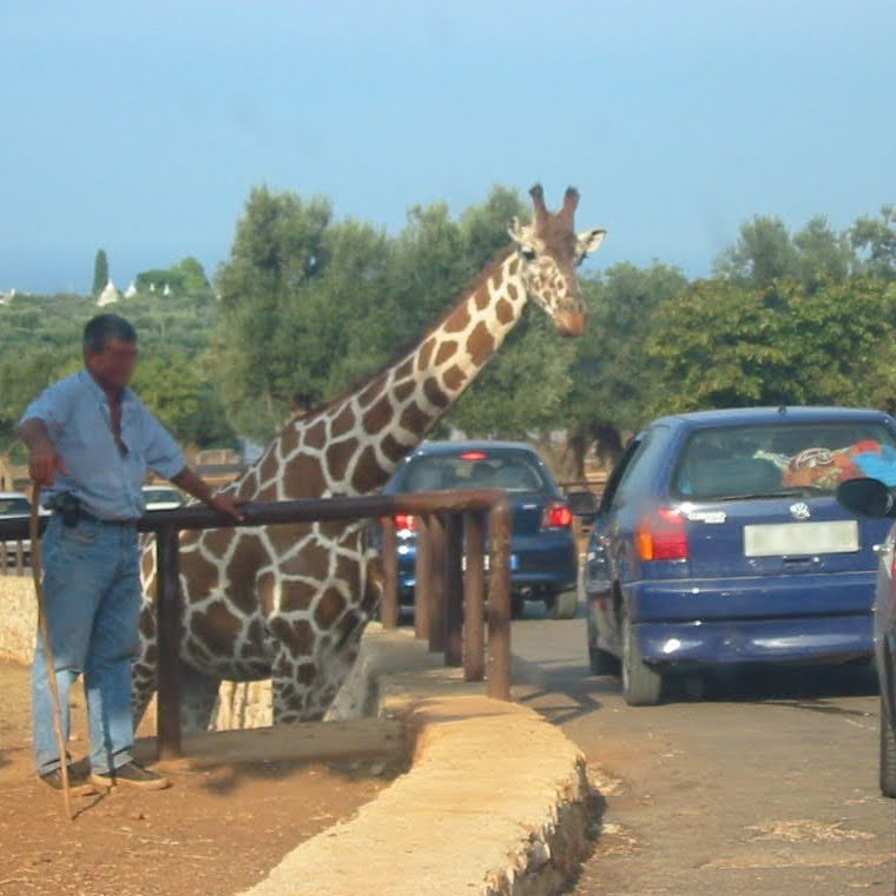}
  \end{minipage}\hfill
  \begin{minipage}[t]{.34\linewidth}
    \centering
    \includegraphics[width=1\linewidth]{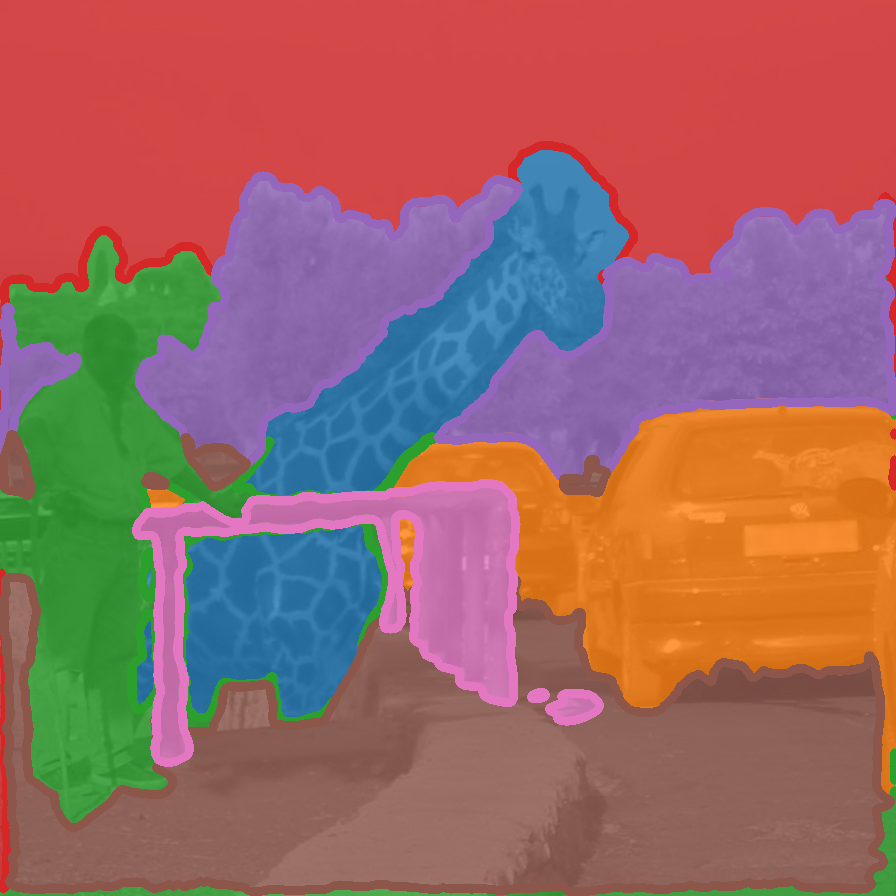}
  \end{minipage}\hfill
  \begin{minipage}[t]{.30\linewidth}
    \centering
    \small
    \begin{tabular}[b]{lll}
        \toprule
        & Color & Name \\
        \midrule
& \cellcolor[HTML]{1f77b4} & Tall giraffe \\
& \cellcolor[HTML]{ff7f0e} & Blue automobile \\
& \cellcolor[HTML]{2ca02c} & Tanned man in shirt and pants \\
& \cellcolor[HTML]{d62728} & Open sky \\
& \cellcolor[HTML]{9467bd} & Trees, bushes \\
& \cellcolor[HTML]{8c564b} & Dirt road, sandy ground \\
& \cellcolor[HTML]{e377c2} & Wood railing, fence \\
        \bottomrule
    \end{tabular}
  \end{minipage}
  \begin{minipage}[t]{.34\linewidth}
    \centering
    \includegraphics[width=1\linewidth]{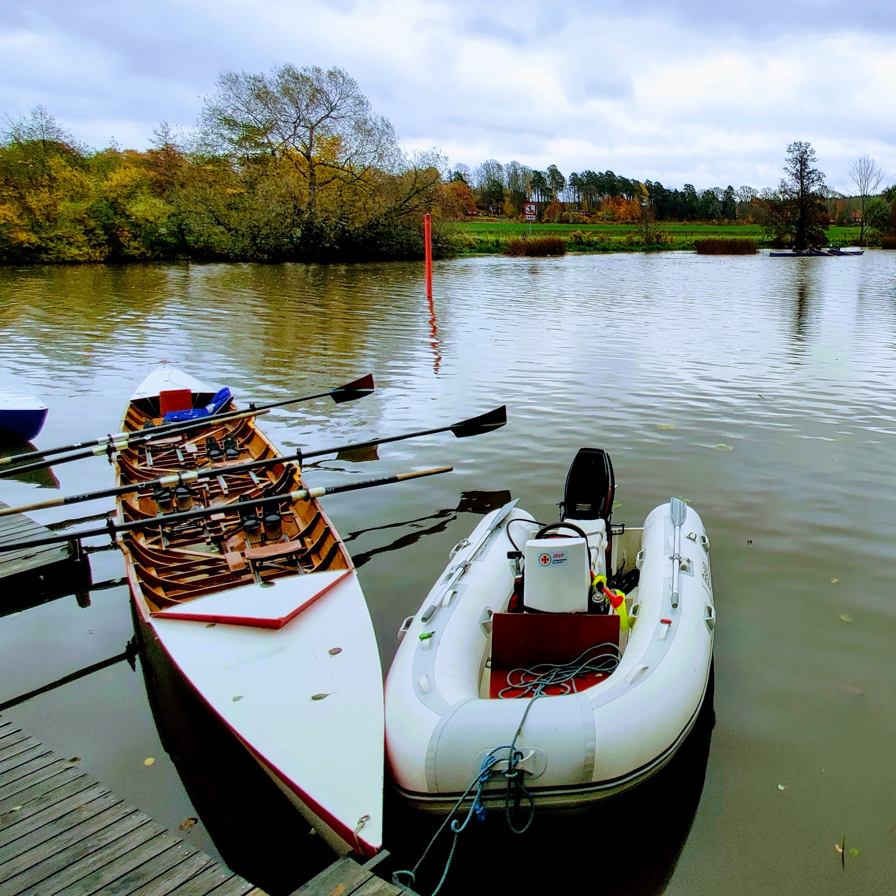}
  \end{minipage}\hfill
  \begin{minipage}[t]{.34\linewidth}
    \centering
    \includegraphics[width=1\linewidth]{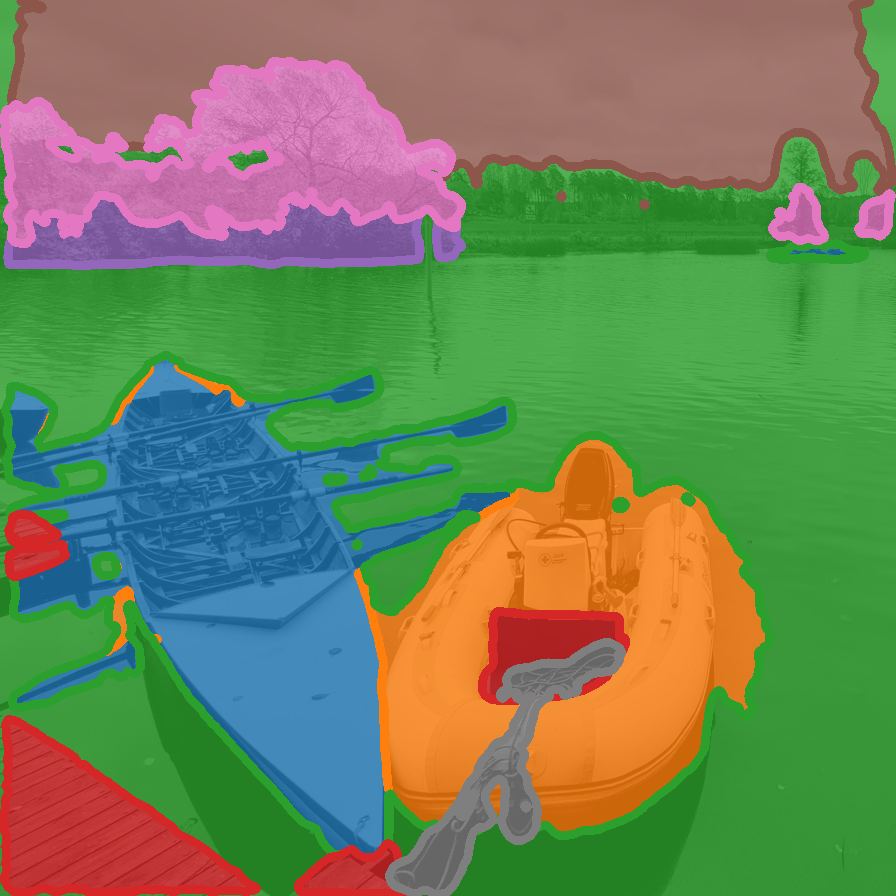}
  \end{minipage}\hfill
  \begin{minipage}[t]{.30\linewidth}
    \centering
    \small
    \begin{tabular}[b]{lll}
        \toprule
        & Color & Name \\
        \midrule
& \cellcolor[HTML]{1f77b4} & Wood rowing canoe \\
& \cellcolor[HTML]{ff7f0e} & Inflatable motor boat \\
& \cellcolor[HTML]{2ca02c} & Peaceful lake \\
& \cellcolor[HTML]{d62728} & Wooden pier \\
& \cellcolor[HTML]{9467bd} & Bush \\
& \cellcolor[HTML]{8c564b} & Blue sky \\
& \cellcolor[HTML]{e377c2} & Tree \\
& \cellcolor[HTML]{7f7f7f} & Nautical rope \\
        \bottomrule
    \end{tabular}
  \end{minipage}
  \begin{minipage}[t]{.34\linewidth}
    \centering
    \includegraphics[width=1\linewidth]{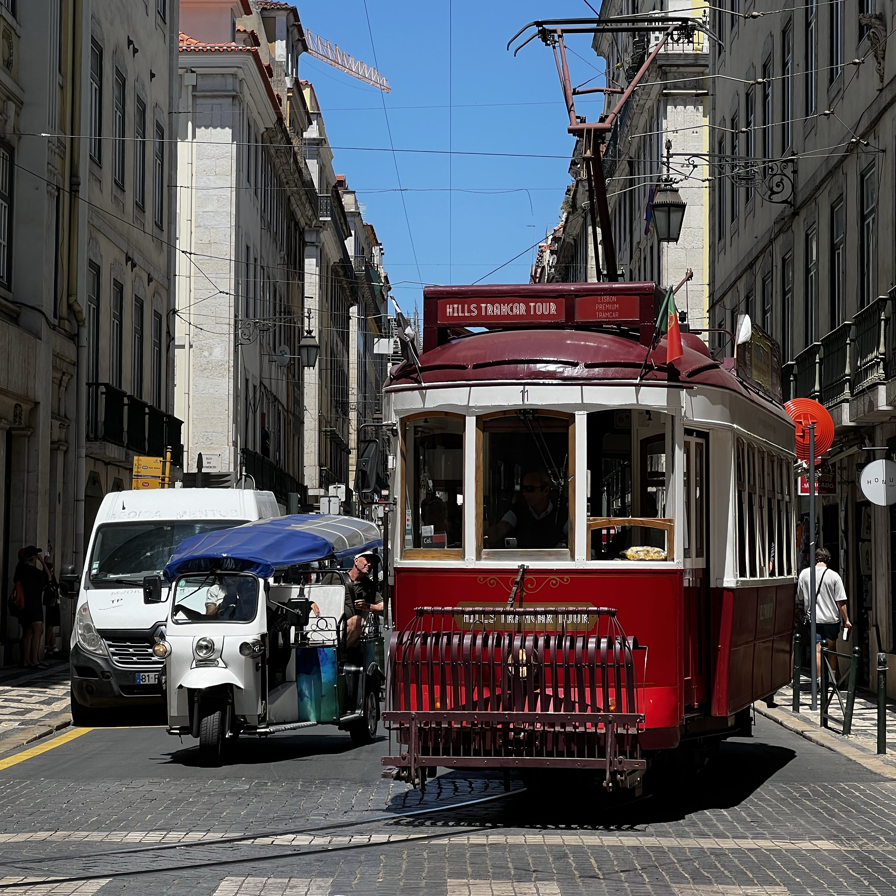}
  \end{minipage}\hfill
  \begin{minipage}[t]{.34\linewidth}
    \centering
    \includegraphics[width=1\linewidth]{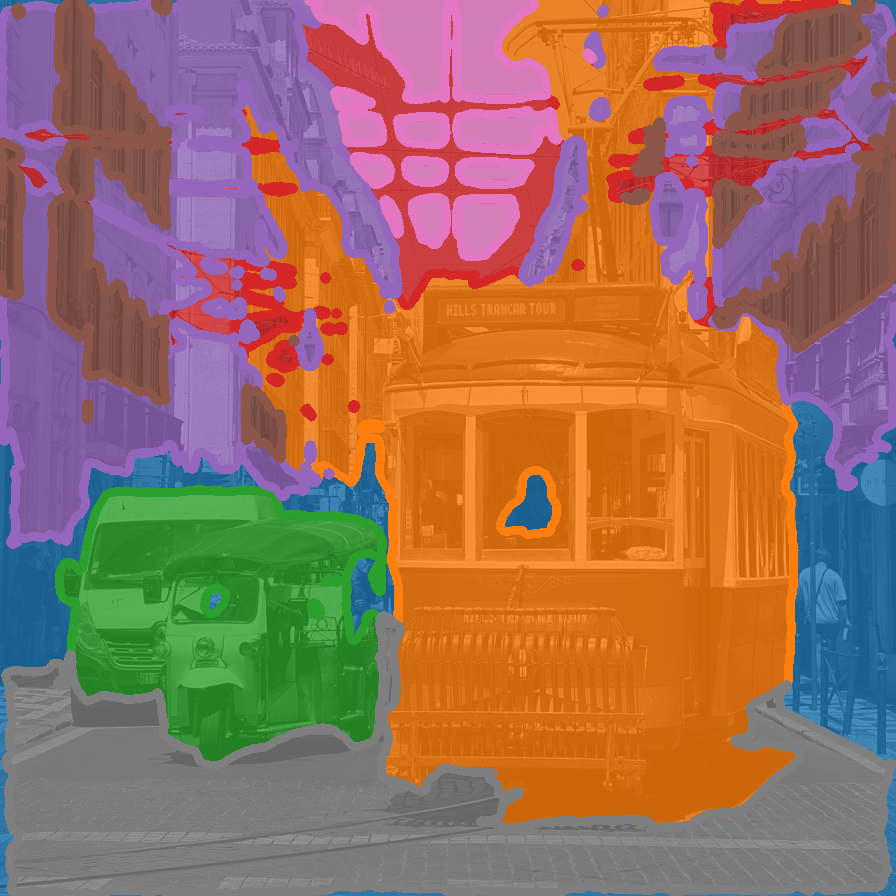}
  \end{minipage}\hfill
  \begin{minipage}[t]{.30\linewidth}
    \centering
    \small
    \begin{tabular}[b]{lll}
        \toprule
        & Color & Name \\
        \midrule
        & \cellcolor[HTML]{1f77b4} & Pedestrian \\
        & \cellcolor[HTML]{ff7f0e} & Tram \\
        & \cellcolor[HTML]{2ca02c} & Car \\
        & \cellcolor[HTML]{d62728} & Electric wires \\
        & \cellcolor[HTML]{9467bd} & Facade \\
        & \cellcolor[HTML]{8c564b} & Window \\
        & \cellcolor[HTML]{e377c2} & Open sky \\
        & \cellcolor[HTML]{7f7f7f} & Road, pavement \\
        \bottomrule
    \end{tabular}
  \end{minipage}
  \caption{\textbf{Open-vocabulary semantic segmentation, part 2/2.} The input resolution is $896{\times}896$ pixels.}
  \label{fig:supmat-open-vocab-segmentation-part-1}
\end{figure*}

\paragraph{Open-vocabulary semantic segmentation.}
Figures~\ref{fig:supmat-open-vocab-segmentation-part-2}-\ref{fig:supmat-open-vocab-segmentation-part-1} demonstrate that the segmentation results of \ours with images and texts in the wild.
For each image, we select a small number of descriptive text prompts and run the zero-shot semantic segmentation pipeline described in \autoref{sec:zero-shot-classification-retrieval-segmentation}.
Our model is able to segment complex scenes with multiple semantic objects and specific text inputs, \eg ``pesto bruschetta'' and ``nautical rope''.

\begin{landscape}
\begin{table}
    \resizebox{1.0\linewidth}{!}{
    \begin{tabular}{@{}ll c ll@{}}
    \toprule
    Original & Optimized && Original & Optimized \\
    \midrule
wall & wall && swivel chair & swivel chair \\
building, edifice & \bfseries facade, frontage, frontal && boat & boat \\
sky & sky && bar & bar \\
floor, flooring & \bfseries floor && arcade machine & arcade machine \\
tree & tree && hovel, hut, hutch, shack, shanty & \bfseries hovel \\
ceiling & ceiling && bus, autobus, coach, charabanc, double-decker, jitney, motorbus, motorcoach, omnibus, passenger vehicle & \bfseries bus \\
road, route & \bfseries road && towel & towel \\
bed  & \bfseries bed && light, light source & \bfseries skylight, fanlight \\
windowpane, window  & \bfseries windowpane && truck, motortruck & \bfseries truck \\
grass & grass && tower & tower \\
cabinet & cabinet && chandelier, pendant, pendent & \bfseries chandelier \\
sidewalk, pavement & sidewalk, pavement && awning, sunshade, sunblind & \bfseries awning \\
person, individual, someone, somebody, mortal, soul & \bfseries people && streetlight, street lamp & \bfseries streetlight \\
earth, ground & \bfseries ground, earth && booth, cubicle, stall, kiosk & \bfseries newsstand \\
door, double door & \bfseries interior door && television receiver, television, television set, tv, tv set, idiot box, boob tube, telly, goggle box & \bfseries television receiver \\
table & table && airplane, aeroplane, plane & \bfseries airplane \\
mountain, mount & \bfseries mountain && dirt track & dirt track \\
plant, flora, plant life & \bfseries bush && apparel, wearing apparel, dress, clothes & \bfseries clothes closet, clothespress \\
curtain, drape, drapery, mantle, pall & \bfseries curtain && pole & pole \\
chair & chair && land, ground, soil & \bfseries land \\
car, auto, automobile, machine, motorcar & \bfseries car && bannister, banister, balustrade, balusters, handrail & bannister, banister, balustrade, balusters, handrail \\
water & water && escalator, moving staircase, moving stairway & \bfseries escalator \\
painting, picture & \bfseries painting && ottoman, pouf, pouffe, puff, hassock & \bfseries footstool, footrest, ottoman, tuffet \\
sofa, couch, lounge & sofa, couch, lounge && bottle & bottle \\
shelf & shelf && buffet, counter, sideboard & \bfseries china cabinet, china closet \\
house & house && poster, posting, placard, notice, bill, card & \bfseries poster \\
sea & sea && stage & stage \\
mirror & mirror && van & van \\
rug, carpet, carpeting & \bfseries rug && ship & ship \\
field & field && fountain & fountain \\
armchair & armchair && conveyer belt, conveyor belt, conveyer, conveyor, transporter & \bfseries conveyer belt \\
seat & seat && canopy & \bfseries baldachin \\
fence, fencing & \bfseries fence && washer, automatic washer, washing machine & \bfseries washer \\
desk & desk && plaything, toy & \bfseries plaything \\
rock, stone & \bfseries rock && swimming pool, swimming bath, natatorium & \bfseries swimming pool \\
wardrobe, closet, press & \bfseries wardrobe && stool & stool \\
lamp & lamp && barrel, cask & \bfseries barrel \\
bathtub, bathing tub, bath, tub & \bfseries bathtub && basket, handbasket & \bfseries basket \\
railing, rail & \bfseries railing && waterfall, falls & \bfseries waterfall \\
cushion & \bfseries pillow && tent, collapsible shelter & \bfseries tent \\
base, pedestal, stand & \bfseries stall, stand, sales booth && bag & bag \\
box & box && minibike, motorbike & \bfseries motorcycle, bike \\
column, pillar & \bfseries column && cradle & \bfseries baby bed, baby's bed \\
signboard, sign & \bfseries signboard && oven & oven \\
chest of drawers, chest, bureau, dresser & \bfseries chest of drawers && ball & ball \\
counter & \bfseries reception desk && food, solid food & \bfseries food \\
sand & sand && step, stair & \bfseries pedestal, plinth, footstall \\
sink & sink && tank, storage tank & \bfseries tank \\
skyscraper & skyscraper && trade name, brand name, brand, marque & \bfseries trade name \\
fireplace, hearth, open fireplace & fireplace, hearth, open fireplace && microwave, microwave oven & \bfseries microwave \\
refrigerator, icebox & \bfseries refrigerator && pot, flowerpot & \bfseries pot \\
grandstand, covered stand & \bfseries grandstand && animal, animate being, beast, brute, creature, fauna & \bfseries animal \\
path & path && bicycle, bike, wheel, cycle  & \bfseries bicycle \\
stairs, steps & \bfseries stairs && lake & lake \\
runway & runway && dishwasher, dish washer, dishwashing machine & \bfseries dishwasher \\
case, display case, showcase, vitrine & case, display case, showcase, vitrine && screen, silver screen, projection screen & \bfseries screen \\
pool table, billiard table, snooker table & \bfseries pool table && blanket, cover & \bfseries blanket \\
pillow & \bfseries pillow sham && sculpture & sculpture \\
screen door, screen & \bfseries shower && hood, exhaust hood & \bfseries range hood \\
stairway, staircase & \bfseries stairway && sconce & sconce \\
river & river && vase & vase \\
bridge, span & \bfseries bridge && traffic light, traffic signal, stoplight & \bfseries traffic light \\
bookcase & bookcase && tray & tray \\
blind, screen & \bfseries blind && ashcan, trash can, garbage can, wastebin, ash bin, ash-bin, ashbin, dustbin, trash barrel, trash bin & ashcan, trash can, garbage can, wastebin, ash bin, ash-bin, ashbin, dustbin, trash barrel, trash bin \\
coffee table, cocktail table & \bfseries coffee table && fan & fan \\
toilet, can, commode, crapper, pot, potty, stool, throne & \bfseries toilet && pier, wharf, wharfage, dock & \bfseries pier \\
flower & flower && crt screen & crt screen \\
book & book && plate & \bfseries plate, collection plate \\
hill & \bfseries hillside && monitor, monitoring device & \bfseries computer screen, computer display \\
bench & bench && bulletin board, notice board & \bfseries bulletin board \\
countertop & countertop && shower & shower \\
stove, kitchen stove, range, kitchen range, cooking stove & stove, kitchen stove, range, kitchen range, cooking stove && radiator & radiator \\
palm, palm tree & \bfseries cabbage palm, cabbage tree, Livistona australis && glass, drinking glass & \bfseries glass \\
kitchen island & kitchen island && clock & clock \\
computer, computing machine, computing device, data processor, electronic computer, information processing system & \bfseries desktop computer && flag & flag \\
    \bottomrule
    \end{tabular}
    }
    \caption{\textbf{ADE20K dataset:} original class names \vs optimized class names for zero-shot semantic segmentation. Modified class names are highlighted in bold. The new class names have been picked through manual analysis to increase specificity, for example \textit{building} to \textit{facade}, or to remove potential confusion, for example \textit{throne} for \textit{toilet}.}
    \label{supmat:classnames}
\end{table}
\end{landscape}
\clearpage

